\pdfoutput=1
\documentclass[11pt]{article}
\usepackage[preprint]{acl}
\usepackage{times}
\usepackage{latexsym}
\usepackage[T1]{fontenc}
\usepackage[utf8]{inputenc}
\usepackage{microtype}
\usepackage{inconsolata}
\usepackage{graphicx}
\usepackage{longtable}
\usepackage{booktabs}
\usepackage{multirow}
\usepackage{colortbl}
\usepackage{tikz}
\usepackage{fvextra}
\usepackage{verbatim}
\usepackage{enumitem}
\usepackage{listings}
\usepackage{xcolor}
\usepackage{float}
\usepackage{caption} 

\title{Visuospatial Cognitive Assistant\thanks{This is a technical report and a draft version. Work in progress.}}

\author{
Qi Feng${}^{1}$\\
${}^{1}$Kyoto University\\
\texttt{feng.qi.45n@st.kyoto-u.ac.jp}\\
\url{https://huggingface.co/nkkbr/ViCA}
}

\begin{document}
\maketitle

\begin{abstract}

Video-based spatial cognition is vital for robotics and embodied AI but challenges current Vision-Language Models (VLMs). This paper makes two key contributions. First, we introduce \textbf{ViCA\footnote{\textbf{ViCA}: \textbf{Vi}suospatial \textbf{C}ognitive \textbf{A}ssistant}-322K}, a diverse dataset of 322,003 QA pairs from real-world indoor videos (ARKitScenes, ScanNet, ScanNet++), offering supervision for 3D metadata-grounded queries and video-based complex reasoning. Second, we develop \textbf{ViCA-7B}, fine-tuned on ViCA-322K, which achieves new state-of-the-art on all eight VSI-Bench tasks, outperforming existing models, including larger ones (e.g., +26.1 on Absolute Distance). For interpretability, we present \textbf{ViCA-Thinking-2.68K}, a dataset with explicit reasoning chains, and fine-tune \textbf{ViCA-7B} to create \textbf{ViCA-7B-Thinking}, a model that articulates its spatial reasoning. Our work highlights the importance of targeted data and suggests paths for improved temporal-spatial modeling. We release all resources to foster research in robust visuospatial intelligence.

\end{abstract}

\section{Introduction}

Large Vision-Language Models (VLMs) have demonstrated remarkable proficiency in multimodal understanding and generation \citep{bai2025qwen2, team2024gemini, liu2023visual, zhu2025internvl3, wang2024emu3}, advancing tasks like visual question answering and image captioning. However, a critical frontier, particularly for robotics, augmented reality, and embodied AI, is sophisticated video-based spatial cognition—the ability to perceive, reason about, and interact with the three-dimensional (3D) structure and dynamics of environments from video.

\begin{figure}[t]
  \includegraphics[width=\columnwidth]{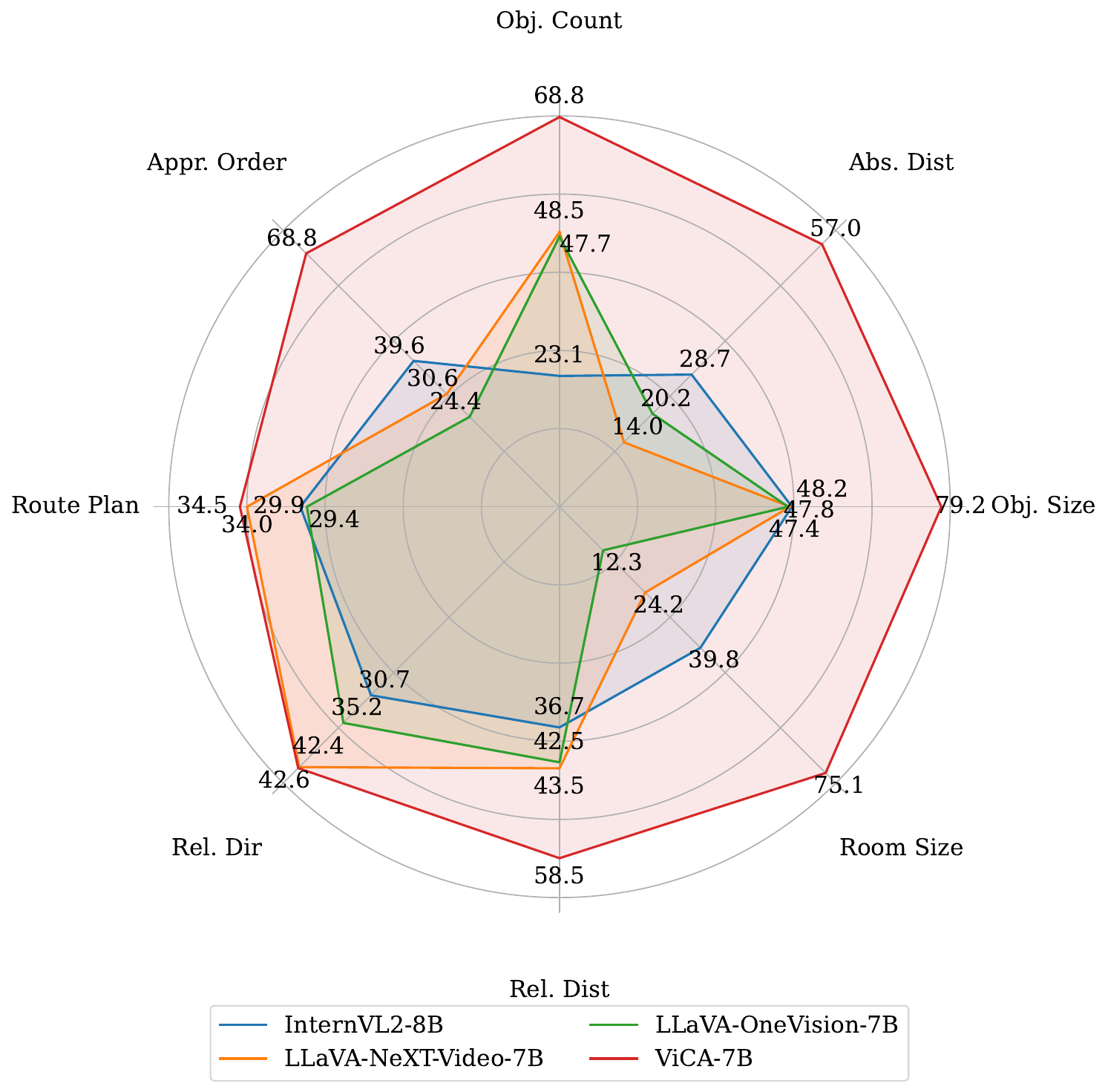}
  \caption{\textbf{Performance comparison of ViCA-7B and existing 7B/8B-scale models across eight spatial reasoning tasks in VSI-Bench.} ViCA-7B consistently outperforms all baselines, demonstrating superior visuospatial understanding capabilities.}
  \label{fig:first_img}
\end{figure}

Current VLMs often struggle with tasks demanding nuanced spatial understanding beyond simple object recognition. Challenges include comprehending fine-grained spatial relationships, tracking appearance and geometric changes, precisely estimating sizes and distances, and reasoning about navigability or functional affordances. This gap is partly due to a scarcity of large-scale, diverse datasets offering rich, targeted supervision for these intricate spatial skills. Existing datasets LLaVA-Instruct-150K\citep{liu2023visual}, COCO Captions\citep{chen2015microsoft}, MINT-1T\citep{awadalla2024mint}, OmniCorpus\citep{li2024omnicorpus} and Infinity-MM\citep{gu2024infinity} may focus on static images, offer limited spatial query diversity, or lack the precise 3D ground truth crucial for robust geometric understanding. Consequently, models may fail to generalize, and their internal reasoning processes often remain opaque, hindering trust in critical applications.

To advance video-based spatial cognition, we argue for a concerted effort in two key directions: (1) developing comprehensive datasets that cover a wide spectrum of spatial reasoning tasks, grounded in rich, real-world video data; and (2) creating models that can effectively leverage such data for superior performance while offering insights into their reasoning pathways.

In this work, we address these challenges. Our contributions can be summarized as follows:
\begin{itemize}
\item We introduce \textbf{ViCA-322K}\footnote{Available at \url{https://huggingface.co/datasets/nkkbr/ViCA-322K}}, a large-scale, diverse dataset for video-based spatial cognition, containing 322,003 high-quality question-answer pairs. Derived from three real-world indoor video datasets (ARKitScenes, ScanNet, ScanNet++), it spans direct spatial queries grounded in 3D metadata and complex reasoning tasks based on video observations.
\item We develop \textbf{ViCA-7B}by fine-tuning a state-of-the-art VLM on ViCA-322K. ViCA-7B achieves new state-of-the-art performance, comprehensively and significantly outperforming current closed-source and open-source models across all eight tasks on the VSI-Bench benchmark, notably surpassing larger 72B open-source models with its 7B parameter scale.
\item We construct \textbf{ViCA-Thinking-2.68K}\footnote{Available at \url{https://huggingface.co/datasets/nkkbr/ViCA-thinking-2.68k}}, a novel dataset incorporating explicit reasoning chains for spatial cognitive tasks. Using this, we fine-tune ViCA-7B to create \textbf{ViCA-7B-Thinking}, a model that can articulate its step-by-step reasoning process, thereby enhancing model interpretability.
\item We publicly release our ViCA-7B and ViCA-7B-Thinking models, evaluation code, results, fine-tuning scripts, and training logs, aiming to foster further research, ensure reproducibility, and contribute to the broader community's efforts in advancing visuospatial intelligence.
\end{itemize}

\section{Related Work}

\paragraph{Large Vision-Language Models (VLMs)}
The landscape of AI has been significantly reshaped by the advent of Large Language Models (LLMs) and their subsequent extension to multimodal domains, giving rise to VLMs. Models like GPT-4V \citep{openai2023gpt4v}, Gemini \citep{gemini2023report}, LLaVA \citep{liu2023visual}, Flamingo \citep{alayrac2022flamingo}, and BLIP \citep{li2022blip, li2023blip, xue2024xgen} have demonstrated impressive abilities in zero-shot and few-shot learning across a variety of vision-language tasks. Many recent efforts have focused on extending these capabilities to video understanding, with models such as Video-LLaMA \citep{zhang2023video}, VideoChat \citep{li2023videochat}, and LLaVA-NeXT-Video \citep{zhang2024llavanextvideo} aiming to process and reason about temporal sequences. While these models exhibit strong general visual understanding, our work focuses on enhancing their proficiency in a specific, challenging domain: nuanced video-based spatial cognition, which often requires more than recognizing objects and actions over time.

\paragraph{Models and Benchmarks for Spatial Cognition}
Reasoning about 3D space from visual input is a long-standing challenge in computer vision and robotics. While traditional approaches often relied on explicit 3D reconstruction \citep{kerbl20233d} and geometric algorithms \citep{pautrat2023gluestick}, recent VLM-based methods aim for end-to-end spatial understanding. The VSI-Bench \citep{yang2024thinking}, which we use for evaluation, represents a significant step towards standardized assessment of such capabilities in real-world video settings. Prior benchmarks Video-mme \citep{fu2024video}, Egoschema \citep{mangalam2023egoschema}, Mvbench \citep{li2024mvbench} and Video-Bench \citep{ning2023video} might touch upon spatial aspects (e.g., object localization within an image or relative positioning), but VSI-Bench specifically targets a comprehensive suite of video-based spatial memory and reasoning tasks. Our ViCA-7B model, fine-tuned on ViCA-322K, aims to directly address the types of sophisticated spatial reasoning VSI-Bench evaluates, demonstrating how targeted data can significantly enhance VLM performance in this domain.

\section{Dataset: ViCA-322K}

\begin{figure*}[!t]
  \centering
  \begin{minipage}{.4\textwidth}
  \includegraphics[width=\linewidth]{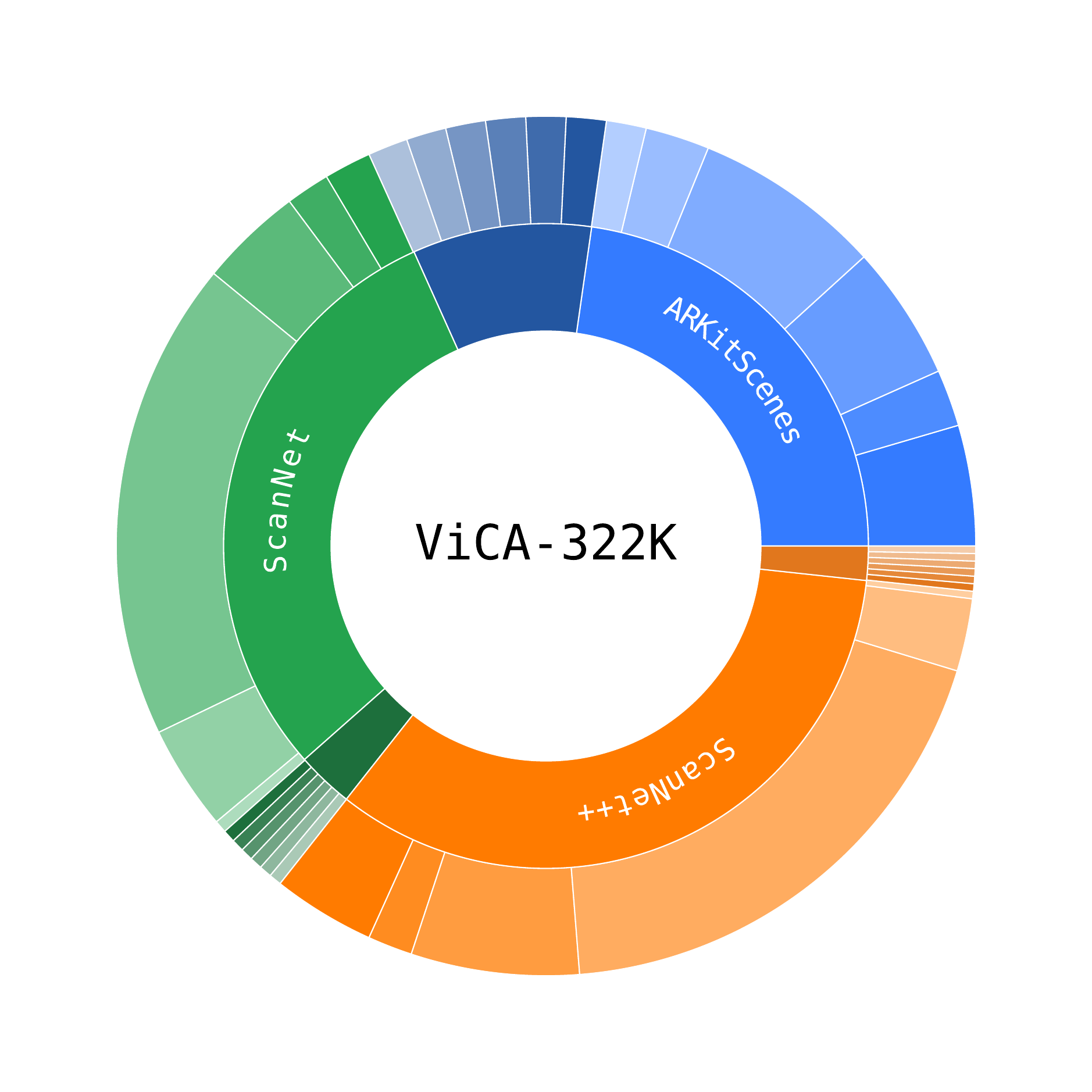}
  \end{minipage}%
  \hfill
  \begin{minipage}{.56\textwidth}
    \scriptsize
    \renewcommand{\arraystretch}{1.1}
    \setlength\tabcolsep{4pt}
    
    \begin{minipage}{\textwidth}
    \centering
    \textbf{ArkitScenes}
    \vspace{0.5em}
    
    \begin{minipage}[b]{.48\textwidth}
    \begin{tabular}{l l r}
    \tikz\fill[fill={rgb,255:red,52; green,123; blue,255}] (0,0) rectangle (0.8em,0.8em); & object count & 13616 \\
    \tikz\fill[fill={rgb,255:red,77; green,140; blue,255}] (0,0) rectangle (0.8em,0.8em); & object relative distance & 6416 \\
    \tikz\fill[fill={rgb,255:red,103; green,156; blue,255}] (0,0) rectangle (0.8em,0.8em); & object size estimation & 15217 \\
    \tikz\fill[fill={rgb,255:red,128; green,172; blue,255}] (0,0) rectangle (0.8em,0.8em); & object abs distance & 21302 \\
    \tikz\fill[fill={rgb,255:red,154; green,189; blue,255}] (0,0) rectangle (0.8em,0.8em); & obj appearance order & 7230 \\
    \tikz\fill[fill={rgb,255:red,179; green,206; blue,255}] (0,0) rectangle (0.8em,0.8em); & room size & 4494 \\
    \end{tabular}
    \end{minipage}
    \hfill
    \begin{minipage}[b]{.48\textwidth}
    \begin{tabular}{l l r}
    \tikz\fill[fill={rgb,255:red,35; green,86; blue,160}] (0,0) rectangle (0.8em,0.8em); & conversation & 4495 \\
    \tikz\fill[fill={rgb,255:red,63; green,107; blue,172}] (0,0) rectangle (0.8em,0.8em); & important daily necessities & 4495 \\
    \tikz\fill[fill={rgb,255:red,90; green,128; blue,184}] (0,0) rectangle (0.8em,0.8em); & furniture & 4495 \\
    \tikz\fill[fill={rgb,255:red,118; green,149; blue,196}] (0,0) rectangle (0.8em,0.8em); & usage & 4495 \\
    \tikz\fill[fill={rgb,255:red,145; green,171; blue,208}] (0,0) rectangle (0.8em,0.8em); & wheelchair user & 4495 \\
    \tikz\fill[fill={rgb,255:red,172; green,192; blue,219}] (0,0) rectangle (0.8em,0.8em); & spatial description & 4495 \\
    \end{tabular}
    \end{minipage}
    \end{minipage}
    
    \rule{\linewidth}{.4pt}
    
    \begin{minipage}{\textwidth}
    \centering
    \textbf{ScanNet}
    \vspace{0.5em}
    
    \begin{minipage}[b]{.48\textwidth}
    \begin{tabular}{l l r}
    \tikz\fill[fill={rgb,255:red,36; green,163; blue,78}] (0,0) rectangle (0.8em,0.8em); & object count & 5356 \\
    \tikz\fill[fill={rgb,255:red,63; green,174; blue,100}] (0,0) rectangle (0.8em,0.8em); & object relative distance & 4965 \\
    \tikz\fill[fill={rgb,255:red,91; green,186; blue,122}] (0,0) rectangle (0.8em,0.8em); & object size estimation & 11658 \\
    \tikz\fill[fill={rgb,255:red,118; green,197; blue,144}] (0,0) rectangle (0.8em,0.8em); & object abs distance & 54240 \\
    \tikz\fill[fill={rgb,255:red,146; green,209; blue,166}] (0,0) rectangle (0.8em,0.8em); & obj appearance order & 11860 \\
    \tikz\fill[fill={rgb,255:red,173; green,220; blue,189}] (0,0) rectangle (0.8em,0.8em); & room size & 1425 \\
    \end{tabular}
    \end{minipage}%
    \hfill
    \begin{minipage}[b]{.48\textwidth}
    \begin{tabular}{l l r}
    \tikz\fill[fill={rgb,255:red,29; green,111; blue,60}] (0,0) rectangle (0.8em,0.8em); & conversation & 1425 \\
    \tikz\fill[fill={rgb,255:red,57; green,129; blue,84}] (0,0) rectangle (0.8em,0.8em); & important daily necessities & 1425 \\
    \tikz\fill[fill={rgb,255:red,86; green,147; blue,109}] (0,0) rectangle (0.8em,0.8em); & furniture & 1425 \\
    \tikz\fill[fill={rgb,255:red,114; green,165; blue,133}] (0,0) rectangle (0.8em,0.8em); & usage & 1425 \\
    \tikz\fill[fill={rgb,255:red,142; green,183; blue,158}] (0,0) rectangle (0.8em,0.8em); & wheelchair user & 1425 \\
    \tikz\fill[fill={rgb,255:red,170; green,201; blue,182}] (0,0) rectangle (0.8em,0.8em); & spatial description & 1425 \\
    \end{tabular}
    \end{minipage}
    \end{minipage}
    
    \rule{\linewidth}{.4pt}
    
    \begin{minipage}{\textwidth}
    \centering
    \textbf{ScanNet++}
    \vspace{0.5em}
    
    \begin{minipage}[b]{.48\textwidth}
    \begin{tabular}{l l r}
    \tikz\fill[fill={rgb,255:red,255; green,123; blue,0}] (0,0) rectangle (0.8em,0.8em); & object count & 11580 \\
    \tikz\fill[fill={rgb,255:red,255; green,140; blue,32}] (0,0) rectangle (0.8em,0.8em); & object relative distance & 5052 \\
    \tikz\fill[fill={rgb,255:red,255; green,156; blue,64}] (0,0) rectangle (0.8em,0.8em); & object size estimation & 18968 \\
    \tikz\fill[fill={rgb,255:red,255; green,172; blue,96}] (0,0) rectangle (0.8em,0.8em); & object abs distance & 57219 \\
    \tikz\fill[fill={rgb,255:red,255; green,189; blue,128}] (0,0) rectangle (0.8em,0.8em); & obj appearance order & 8200 \\
    \tikz\fill[fill={rgb,255:red,255; green,206; blue,159}] (0,0) rectangle (0.8em,0.8em); & room size & 854 \\
    \end{tabular}
    \end{minipage}%
    \hfill
    \begin{minipage}[b]{.48\textwidth}
    \begin{tabular}{l l r}
    \tikz\fill[fill={rgb,255:red,225; green,119; blue,29}] (0,0) rectangle (0.8em,0.8em); & conversation & 854 \\
    \tikz\fill[fill={rgb,255:red,229; green,136; blue,57}] (0,0) rectangle (0.8em,0.8em); & important daily necessities & 854 \\
    \tikz\fill[fill={rgb,255:red,232; green,153; blue,86}] (0,0) rectangle (0.8em,0.8em); & furniture & 854 \\
    \tikz\fill[fill={rgb,255:red,236; green,170; blue,114}] (0,0) rectangle (0.8em,0.8em); & usage & 854 \\
    \tikz\fill[fill={rgb,255:red,240; green,187; blue,142}] (0,0) rectangle (0.8em,0.8em); & wheelchair user & 854 \\
    \tikz\fill[fill={rgb,255:red,244; green,204; blue,170}] (0,0) rectangle (0.8em,0.8em); & spatial description & 854 \\
    \end{tabular}
    \end{minipage}
    \end{minipage}
    
    \end{minipage}

\caption{
  \textbf{Hierarchical composition of the ViCA-322K dataset.} The inner ring denotes the proportional distribution across three primary sources—\textit{ARKitScenes}, \textit{ScanNet}, and \textit{ScanNet++}—while the outer ring illustrates the fine-grained breakdown into diverse spatial reasoning task types. This structured diversity is designed to provide broad coverage of visuospatial semantics, offering inductive bias that facilitates robust generalization in multimodal learning.
  }
      
\label{fig:dataset_pie}
\end{figure*}

In this section, we present ViCA-322K, a large-scale dataset specifically designed to foster research in video-based spatial cognition. We begin by detailing the foundational video sources and the overall composition of our dataset. We then describe the two main categories of question-answer pairs: Base Data, grounded in precise 3D metadata, and Complex Spatial Reasoning data, requiring deeper video interpretation. Finally, we provide comprehensive statistics to illustrate the scale and diversity of ViCA-322K(see Figure~\ref{fig:dataset_pie}).

\subsection{Video Sources and Data Composition}
\begin{figure*}[t]
  \centering
  \includegraphics[width=\textwidth]{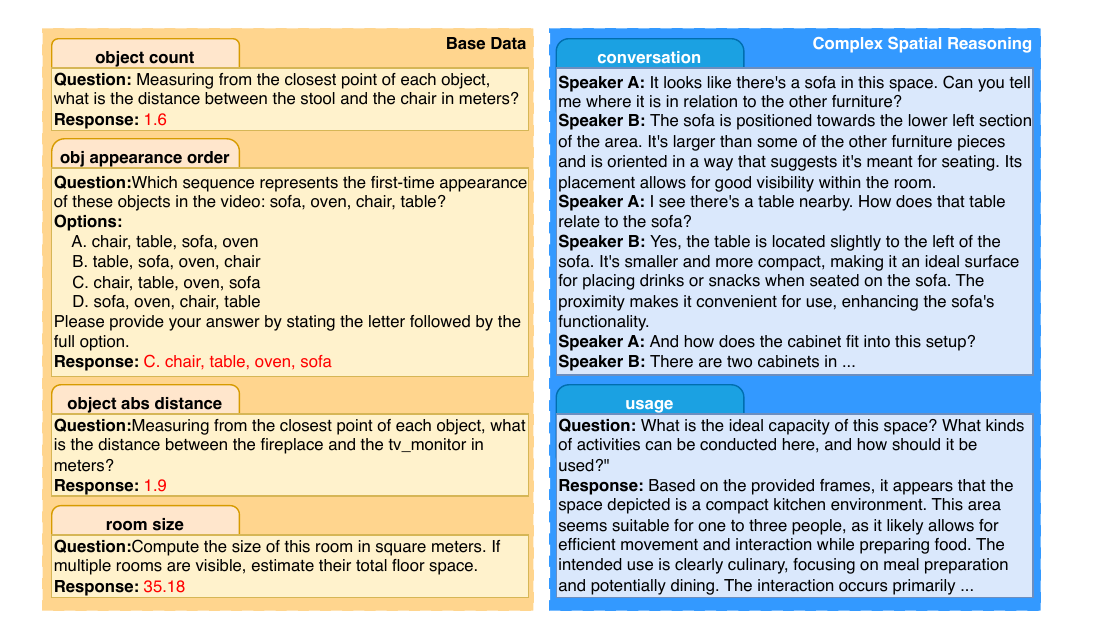}
  \caption{
  \textbf{An illustrative example from our ViCA-322K dataset, showcasing both Base Data and Complex Spatial Reasoning subsets.} The \textbf{Base Data} questions (left) are grounded in 3D metadata such as object counts and distances, allowing for precise supervision. The \textbf{Complex Spatial Reasoning} questions (right) require deeper interpretive understanding from the video alone, including multi-turn spatial dialogues and functional scene understanding. Together, these tasks promote comprehensive evaluation of visuospatial cognitive capabilities.
  }
  \label{fig:vica_sample}
\end{figure*}
The selection of video sources plays a crucial role in video-based spatial understanding. Following the setup of VSI-Bench, we utilize videos from ARKitScenes \citep{baruch2021arkitscenes}, ScanNet \citep{dai2017scannet}, and ScanNet++ \citep{yeshwanth2023scannet++}. We exclusively use the training splits of these datasets and conduct strict filtering to ensure that no videos overlap with those used in VSI-Bench. All videos depict real-world indoor environments, covering a diverse range of room types and layouts, and are accompanied by high-quality annotations.

The resulting dataset can be broadly categorized into two types: (1) data derived from metadata, such as 3D oriented bounding boxes, providing precise supervision; and (2) data requiring complex spatial reasoning based solely on video observations.

\subsection{Base Data}

For the Base Data, we select six spatial reasoning tasks—object count, object relative distance, object size estimation, object absolute distance, object appearance order, and room size—for which precise answers can be generated directly from 3D oriented bounding box annotations.

We deliberately omit two related tasks, object relative direction and route planning, with the objective of encouraging models to transfer learned spatial understanding to novel tasks. This design ensures that the model learns generalizable spatial reasoning abilities, rather than task-specific pattern matching.

\subsection{Complex Spatial Reasoning}

In contrast to Base Data, which is constructed from structured metadata, the Complex Spatial Reasoning subset is designed to probe deeper spatial understanding through natural language questions that require interpreting complex inter-object relations observed in videos.

This subset is further divided into two categories: \textit{general} and \textit{specific}.

The \textit{general} category aims to develop holistic spatial awareness by prompting the model to describe and reason about the overall structure of the environment. In contrast, the \textit{specific} category targets fine-grained spatial reasoning, requiring precise understanding of local object relationships.

By combining these two levels of spatial reasoning, we aim to comprehensively evaluate and enhance both the global and local spatial understanding capabilities of multimodal models.

\subsubsection{General}

The \textit{general} category consists of two types of questions.

First, we construct medium-length multi-turn dialogues between a human and a GPT agent, focusing on discussing the spatial relationships among objects observed in the video. Multi-turn interactions are intended to refine the model's reasoning through iterative clarification and elaboration. During training, the loss is computed only on the GPT agent's responses.

Second, we ask about objects important for daily activities within the scene, prompting the model to consider whether such objects should be repositioned for better visibility or accessibility.

To generate the question-answer pairs, we employ GPT-4o-mini \citep{openai2024gpt4omini}. During generation, only metadata—including object names, coordinates, sizes, and orientations extracted from 3D bounding boxes—is provided. No visual frames or images are supplied.

\subsubsection{Specific}

The \textit{specific} category includes four types of tasks:

\begin{enumerate}

    \item Given a video, determine whether it is possible to add a piece of furniture to the space. If so, specify the type, size, and anticipated improvement to the environment.

    \item Evaluate the number of people the space can accommodate, propose potential usage scenarios, and describe in detail the spatial relationships between users and surrounding objects during usage.

    \item Assess whether the space is accessible to a wheelchair user by choosing one of three options: (i) accessible, (ii) inaccessible, (iii) potentially accessible with minor modifications. Subsequently, provide a detailed description of the user's goals, navigation path, and interactions within the space.

    \item Imagine being situated at a random location within the space and describe the surrounding environment, focusing on the spatial relations between observed objects. To guide the model, we provide three exemplar answers prior to the task.

\end{enumerate}

We generate the specific question-answer pairs using GPT-4o-mini, conditioning on 16 frames uniformly sampled from each video, along with carefully designed prompts.

To encourage linguistic diversity, we design 10 alternative phrasings for each question type and randomly sample one phrasing during the generation process for both the Base Data and Complex Spatial Reasoning subsets.

\subsection{Data Statistics}

The ViCA-322K dataset comprises a total of 322,003 question-answer pairs, collected from ARKitScenes, ScanNet, and ScanNet++. As shown in Figure~\ref{fig:vica_sample}, each dataset contributes both \textbf{Base Data} and \textbf{Complex Spatial Reasoning} samples.\footnote{In ARKitScenes, we include an additional experimental split: Triangular Positional Relationship, \texttt{triangular\_positional\_relationship.json} (21{,}707 entries). Each question asks the model to describe the side lengths and angles of a triangle formed by three specified objects.}

The Base Data subset consists of 281,359 questions derived from structured 3D annotations, covering six spatial reasoning tasks: object count, relative and absolute distances, object size estimation, appearance order, and room size. These questions enable precise, metadata-supervised training.

The Complex Spatial Reasoning subset comprises 40,644 questions, evenly distributed across six functionally grounded categories, including multi-turn conversations, object usage, accessibility assessment, and holistic spatial descriptions. These questions are designed to probe deeper, language-grounded visuospatial understanding based solely on video observations.

Overall, the dataset offers a balanced and diverse benchmark, combining structured spatial perception with rich, context-dependent reasoning. This composition is intended to support the development and evaluation of multimodal models with generalizable visuospatial cognitive capabilities.

\section{Experiments}

This section presents a comprehensive experimental evaluation of our proposed ViCA framework. We first describe the setup for our experiments, detailing the training process for our ViCA models and the VSI-Bench benchmark used for evaluation. We then report the overall performance of ViCA-7B against leading open-source and proprietary models. Following this, we conduct ablation studies to investigate the effects of training data scale and the specific impact of our Complex Spatial Reasoning (CSR) data. Finally, we discuss additional probing experiments to gain deeper insights into model capabilities and current limitations in visuospatial reasoning.

\subsection{Setup}

\subsubsection{Training} We fine-tune the current state-of-the-art open-source vision-language model, LLaVA-Video-7B-Qwen2 \citep{zhang2024video}, using our ViCA-322K dataset. This model is built upon the checkpoint of \texttt{LLaVA-OneVision (SI)}, which has already demonstrated strong video understanding capabilities. To explore data scalability, we trained several variants of the ViCA model on different training splits. The most comprehensive version was fine-tuned on 8 H100 (80GB) GPUs for 55 hours using DeepSpeed ZeRO-3 Offload \citep{rajbhandari2020zero}. We additionally trained ViCA-Base-7B using only the Base portion of ViCA-322K (i.e., excluding the Complex Spatial Reasoning data).

\subsubsection{Evaluation} We evaluate all models on the newly released VSI-Bench, a high-quality benchmark specifically designed to assess multimodal models' visuospatial reasoning and memory capabilities using real-world indoor videos. This benchmark presents challenges that demand not only visual perception and language comprehension, but also reasoning over temporal order, spatial geometry, and egocentric-to-allocentric transformations. 

During evaluation, we strictly follow the prompts provided by VSI-Bench, and do not adopt the \texttt{time\_instruction} feature used in LLaVA-Video examples. Each video is uniformly sampled into 64 frames. In addition, we evaluate two extended variants: one with the inclusion of \texttt{time\_instruction}, and another using a higher frame sampling rate of 128.

\subsection{Overall Results}

\begin{table*}[t]
\small
\centering
\resizebox{\textwidth}{!}{ 
\begin{tabular}{l|c|cccc|cccc}
\toprule
\textbf{Method} & \textbf{Average} & \multicolumn{4}{c|}{\textbf{Numerical Answer}} & \multicolumn{4}{c}{\textbf{Multiple-Choice Answer}} \\
  &  & Obj. Count & Abs. Dist. & Obj. Size & Room Size & Rel. Dist. & Rel. Dir. & Route Plan & Appr. Order \\
\midrule
\rowcolor{gray!20} \multicolumn{10}{l}{\textit{Proprietary Models (API)}} \\
GPT-4o & 34.0 & 46.2 & 5.3 & 43.8 & 38.2 & 37.0 & 41.3 & 31.5 & 28.5 \\
Gemini-1.5 Flash & 42.1 & 49.8 & 30.8 & 53.5 & 54.4 & 37.7 & 41.0 & 31.5 & 37.8 \\
Gemini-1.5 Pro & 45.4 & 56.2 & 30.9 & 64.1 & 43.6 & 51.3 & \cellcolor{gray!40}46.3 & \cellcolor{gray!40}36.0 & 34.6 \\
\midrule
\rowcolor{gray!20} \multicolumn{10}{l}{\textit{Open-source Models}} \\
InternVL2-8B & 34.6 & 23.1 & 28.7 & 48.2 & 39.8 & 36.7 & 30.7 & 29.9 & 39.6 \\
InternVL2-40B & 36.0 & 34.9 & 26.9 & 46.5 & 31.8 & 42.1 & 32.2 & 34.0 & 39.6 \\
VILA-1.5-8B & 28.9 & 17.4 & 21.8 & 50.3 & 18.8 & 32.1 & 34.8 & 31.0 & 24.8 \\
VILA-1.5-40B & 31.2 & 22.4 & 24.8 & 48.7 & 22.7 & 40.5 & 25.7 & 31.5 & 32.9 \\
LLaVA-NeXT-Video-7B & 35.6 & 48.5 & 14.0 & 47.8 & 24.2 & 43.5 & 42.4 & 34.0 & 30.6 \\
LLaVA-NeXT-Video-72B & 40.9 & 48.9 & 22.8 & 57.4 & 35.3 & 42.4 & 36.7 & 35.0 & 48.6 \\
LLaVA-OneVision-7B & 32.4 & 47.7 & 20.2 & 47.4 & 12.3 & 42.5 & 35.2 & 29.4 & 24.4 \\
LLaVA-OneVision-72B & 40.2 & 43.5 & 23.9 & 57.6 & 37.5 & 42.5 & 39.9 & 32.5 & 44.6 \\
ViCA-Base-7B (\textbf{ours}) & 55.4 & 65.6 & 51.3 & 74.9 & 67.1 & 52.0 & 32.6 & 28.4 & \cellcolor{gray!40}\textbf{70.9(+22.3)} \\
ViCA-7B (\textbf{ours}) & \cellcolor{gray!40}\textbf{60.6(+15.2)} & \cellcolor{gray!40}\textbf{68.8(+12.6)} & \cellcolor{gray!40}\textbf{57.0(+26.1)} & \cellcolor{gray!40}\textbf{79.2(+15.1)} & \cellcolor{gray!40}\textbf{75.1(+20.7)} & \cellcolor{gray!40}\textbf{58.5(+7.2)} & \textbf{42.6}(+0.2) & \textbf{34.5}(+0.5) & 68.8 \\
\bottomrule
\end{tabular}
}
\caption{
\textbf{Comparison of different models on VSI-Bench.}
Our ViCA-7B achieves the best performance across most metrics. 
\colorbox{gray!20}{Gray shading} indicates the best overall performance among all models, including 72B-scale and proprietary models, 
while \textbf{bold font} indicates the best performance among open-source models with 7B/8B scale. 
The numbers in parentheses (e.g., +26.1) represent the improvement margins over the best-performing model excluding ViCA-7B and ViCA-Base-7B. 
For \textbf{Rel. Dir.} and \textbf{Route Plan}, the improvements are computed relative to the best scores among other open-source models with 7B/8B scale.
}
\label{tab:comparison}
\end{table*}

Table~\ref{tab:comparison} summarizes the main experimental results. Our model, ViCA-7B, achieves the highest average performance across all eight tasks, outperforming not only open-source models—including larger 72B-scale models—but also several proprietary systems. In the Numerical Answer category, ViCA-7B surpasses all other models. Notably, on the Abs. Dist. task, it outperforms the second-best result by an impressive margin of 26.1 percentage points. This success is largely attributed to our precisely annotated data and the inclusion of Complex Spatial Reasoning (CSR) tasks.

In the Multiple-Choice Answer category, while Gemini-1.5 Pro \citep{deepmind2024gemini15} performs slightly better on Rel. Dir. and Route Plan, our models (ViCA-7B and ViCA-Base-7B) still deliver the best results among all 7B/8B-scale open-source models. Importantly, we did not provide fine-tuning data specific to these two tasks. The observed improvements suggest that the model successfully transfers its learned spatial understanding from other tasks. On the Appr. Order task, ViCA-7B outperforms all other models by over 20 percentage points, which we attribute to our use of YOLO\citep{redmon2016you}-based precise temporal detection for object first appearances during data preparation.

\subsection{Data Scale vs. Performance}

A critical question is: how much data is sufficient for a model to effectively learn these spatial reasoning tasks? Could our data volume be excessive, causing overfitting, or is there still headroom for improvement? To investigate this, we randomly shuffle the training data and save model checkpoints at every 5\% increment of the dataset. We then measure model performance at each checkpoint (see Figure~\ref{fig:csr_scaling}).

The results show that model performance increases significantly from 5\% to 60\% of the training data, indicating that early-stage data plays a pivotal role in building core spatial understanding. However, after surpassing 80\%, performance gains plateau and slightly decrease between 95\% and 100\%. This suggests that our dataset size is well-matched to the model's capacity (7B parameters), and observing performance saturation is crucial to verifying the sufficiency of our data scale relative to model architecture.

\subsection{Effect of Complex Spatial Reasoning Data on Scaling Behavior}

To investigate the impact of our Complex Spatial Reasoning (CSR) data, we conducted a comparative scaling analysis. We trained two main configurations: one using the full ViCA-322K dataset (322K samples, which includes CSR data) and another using only its Base Data subset (281K samples, excluding CSR data). Both models were incrementally fine-tuned, with performance evaluated on VSI-Bench at intervals corresponding to 5\% to 100\% of their respective training datasets.

\begin{figure*}[t]
\centering
\includegraphics[width=\textwidth]{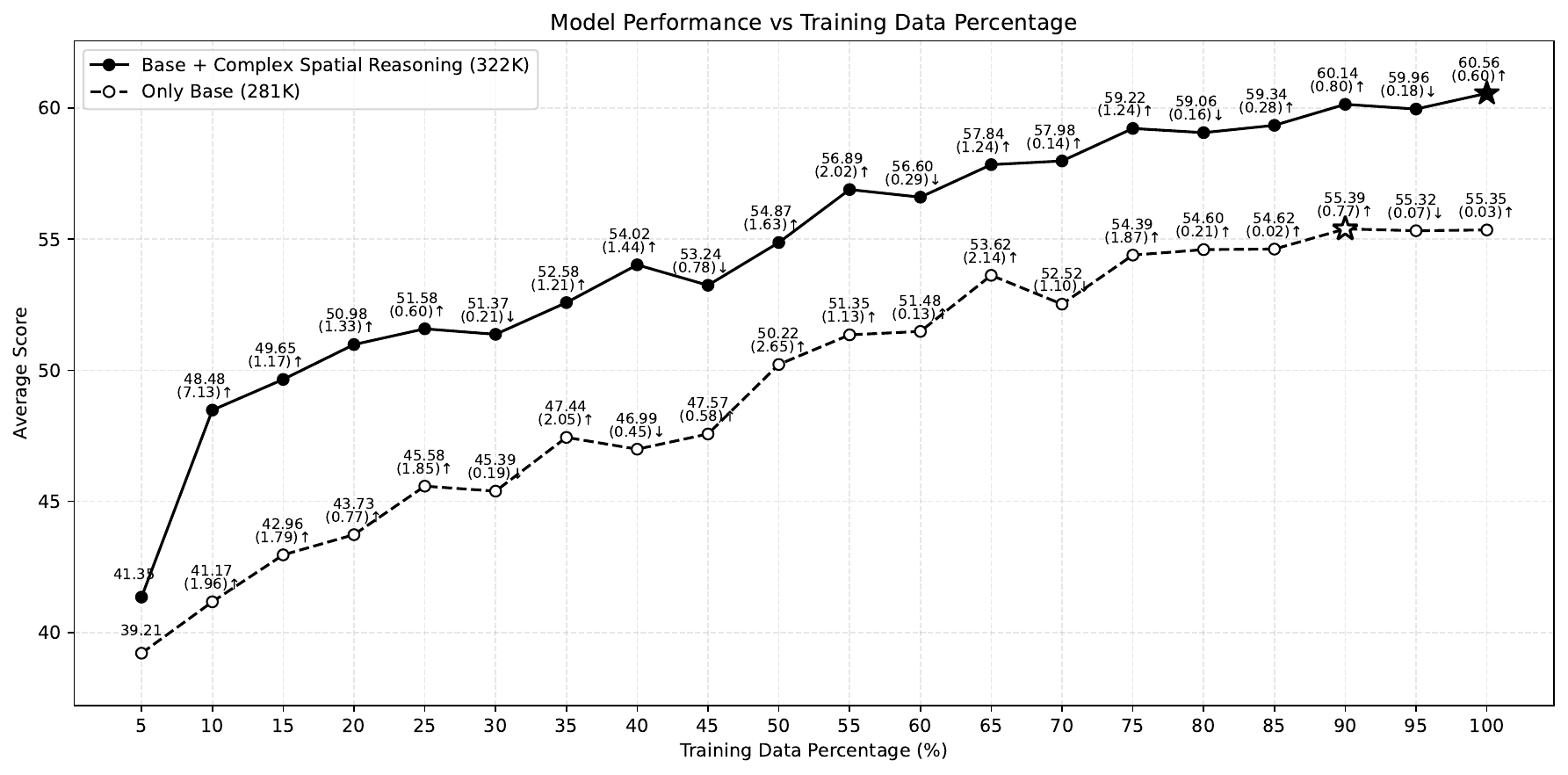}
\caption{
  \textbf{Impact of Complex Spatial Reasoning on model performance scaling.}
  Average benchmark scores with respect to training data percentage.
  The solid line represents the model trained on the full dataset (322K), including Complex Spatial Reasoning data, while the dashed line shows the performance of the model trained only on the base subset (281K).
  The full model consistently outperforms the base-only variant across all scales.
  Score changes between successive checkpoints are annotated, and stars indicate the best performance in each setting.
}
\label{fig:csr_scaling}
\end{figure*}

The results, illustrated in Figure~\ref{fig:csr_scaling}, demonstrate that while both models benefit from increasing amounts of training data, the inclusion of CSR data leads to markedly superior performance. The model trained on the full dataset consistently achieves higher average scores across all data scales, particularly as more training data is utilized (i.e., in higher-resource regimes). Culminating at 100\% data usage, the full model achieves an average score of 60.56 on VSI-Bench, significantly outperforming the base-only model's score of 55.39 by +5.17 points.

These findings strongly suggest that the CSR data component provides crucial inductive biases and richer structural supervision. This, in turn, enhances the model's ability to generalize and tackle complex visuospatial tasks. The substantial performance gap underscores the critical role of diverse and high-quality supervision, such as that provided by our CSR tasks, in advancing spatial cognition capabilities for instruction-tuned models. We therefore advocate for the broader incorporation of such complex reasoning supervision in the development of multimodal datasets to unlock the full potential of large-scale vision-language models.

\subsection{Additional Probing Experiments}
\label{sec:additional_probing}

We further explore whether model performance can be enhanced through means other than large-scale data curation and high-resource fine-tuning.

\subsubsection{Adding Time Instructions to Prompts}

We introduced explicit timing information to the prompt as follows:

\begin{quote}
  \textit{The video lasts for 148.80 seconds, and 64 frames are uniformly sampled from it. These frames are located at 0.00s, 2.33s, 4.70s, ..., 146.40s, 148.77s. Please answer the following questions related to this video.}
\end{quote}

This exposes the model not only to the 64 frames but also to their precise sampling timestamps. Evaluation results (see Appendix Table~\ref{tab:vica-time-instruction}) reveal that including time instructions does not improve performance. In fact, the lengthy \texttt{time\_instruction} text (listing 64 timestamps) slightly degrades model performance. This suggests that the model fails to leverage temporal alignment cues effectively, and instead, the verbosity may interfere with its original reasoning flow.

It is important to note that both LLaVA-Video-7B-Qwen2 and ViCA-7B were trained with such time instructions. Hence, the inability to utilize temporal information reflects a limitation in how well current architectures can extract structured time-based spatial understanding.

\subsubsection{Increasing Input Frames}

We also tested whether providing more visual content helps. While training and inference for LLaVA-Video-7B-Qwen2 and ViCA-7B were consistently conducted using 64 frames, we experimented with increasing the input to 128 frames (see Appendix Table~\ref{tab:vica-frame-comparison}).

Surprisingly, model performance did not improve. The model fails to benefit from the additional visual inputs, likely because it has been conditioned to expect a fixed number of visual tokens. Notably, ViCA-7B was fine-tuned from LLaVA-Video-7B-Qwen2, which itself inherits training from LLaVA-OneVision, exposed to both single/multi-image and video inputs. However, during our training, the number of visual tokens per example was fixed (13440). Increasing to 128 frames doubles the visual token count to 26880, but the model's context window (32768 tokens) still accommodates both input and output sequences. Despite no overflow, performance dropped.

This indicates that simply feeding the model more frames or more detailed temporal metadata does not lead to better spatial understanding. More sophisticated strategies may be required to align temporal information with spatial reasoning effectively.

\section{Probing Model Reasoning via Structured Thought-Response Format}

Although instruction-tuned models have demonstrated strong performance on video-based spatial reasoning tasks, their outputs are typically concise and discrete—either numerical values or multiple-choice letters. While effective for evaluation, such brevity often obscures the model's actual reasoning trajectory and hinders interpretability. To better understand and potentially enhance the model's internal reasoning, we introduce a structured fine-tuning format in which the model must explicitly articulate its reasoning process before providing a final answer.
\subsection{Dataset Construction: Vica-Thinking-2.68K}

We utilize the current state-of-the-art vision-language model, \textit{Gemini 2.5 Pro Preview 03-25}, to construct our \textbf{Vica-Thinking-2.68K} dataset. A distinctive feature of \textit{Gemini 2.5 Pro Preview 03-25} is that, prior to generating a final answer, it internally produces a \textit{Thoughts} section, which contains rich and valuable step-by-step reasoning. However, the Gemini API does not expose these Thoughts to users, limiting direct access to this critical intermediate output.

To address this limitation, we manually collected five high-quality Thoughts samples through the web-based Google AI Studio interface. Specifically, for each video-question pair, we prompted \textit{Gemini 2.5 Pro Preview 03-25} to generate ten candidate answers. From these, we selected the correct answers—either the correct option for multiple-choice questions or answers with a matching rate (MRA) greater than 0.7 for numeric questions. We then extracted the corresponding Thoughts generated during these successful attempts. To ensure quality, we asked ChatGPT-4o to score these Thoughts, and selected the ones with the highest scores for use in our dataset.

In designing the prompting strategy (Appendix~\ref{app:appendix_prompt}), we first instructed the model that it must answer the question based on the given video. We explicitly required that a clear and detailed reasoning process be provided before presenting the final answer. Following this instruction, we included the five high-quality Thoughts examples collected as described above. Additionally, we provided a section titled "Guide for Writing 'Thoughts' (Step-by-Step Reasoning for Video QA)", which offered detailed guidance on structuring and articulating the reasoning process. Finally, we presented the specific question along with a strictly defined output format.

\subsection{Qualitative Analysis of ViCA-7B-Thinking}
\label{sec:qualitative_vica_thinking}

Although fine-tuning for explicit reasoning in \textbf{ViCA-7B-Thinking} resulted in decrease in quantitative task performance compared to ViCA-7B (detailed in Appendix~\ref{app:vica_thinking}), qualitative analysis underscores a significant gain in interpretability. By generating explicit \textit{Thought-Response-Final Answer} outputs, ViCA-7B-Thinking articulates its step-by-step spatial reasoning from video inputs, offering a transparent view into its decision-making process. Representative examples in Appendix~\ref{app:thinking_example} illustrate this capability, showcasing structured and diverse reasoning that adheres to the prescribed format. This enhanced interpretability, despite the accuracy trade-off, is crucial for transparent model evaluation, debugging, and lays groundwork for trustworthy AI in human-in-the-loop applications.

\section{Conclusion}

We introduced ViCA-322K, a large-scale dataset for diverse spatial understanding, from 3D-grounded queries to complex video reasoning. Our ViCA-7B model, fine-tuned on ViCA-322K, achieved state-of-the-art on all VSI-Bench tasks (e.g., +26.1 on Absolute Distance). We found naive input scaling—such as adding detailed time instructions to prompts or increasing input frames—to be ineffective for enhancing spatial reasoning, highlighting the need for structured temporal-spatial modeling. For interpretability, ViCA-Thinking-2.68K and ViCA-7B-Thinking enable articulated reasoning, offering improved explainability despite slightly lower scores. These contributions lay a foundation for advancing spatially-aware multimodal AI.

\section*{Limitations}

Despite strong performance, our work highlights several limitations. Current models, including ViCA-7B, still struggle to effectively utilize explicit temporal information or benefit from increased frame density for spatial reasoning, indicating a need for more sophisticated temporal-spatial modeling techniques. While ViCA-7B-Thinking enhances interpretability by articulating reasoning steps, it currently shows a slight decrease in quantitative task performance compared to its direct-answering counterpart, suggesting a potential trade-off or the need for further refinement in training for explicit reasoning. Finally, the ViCA-322K dataset, while diverse, is derived from indoor environments, and the generation of its complex reasoning components relies on advanced LLMs (GPT-4o-mini, Gemini 2.5 Pro), which may introduce their own biases or limitations into the training data.

\bibliography{custom}

\appendix

\section{Training Records}

\begin{figure*}[p]
  \centering
  \includegraphics[height=0.28\textheight]{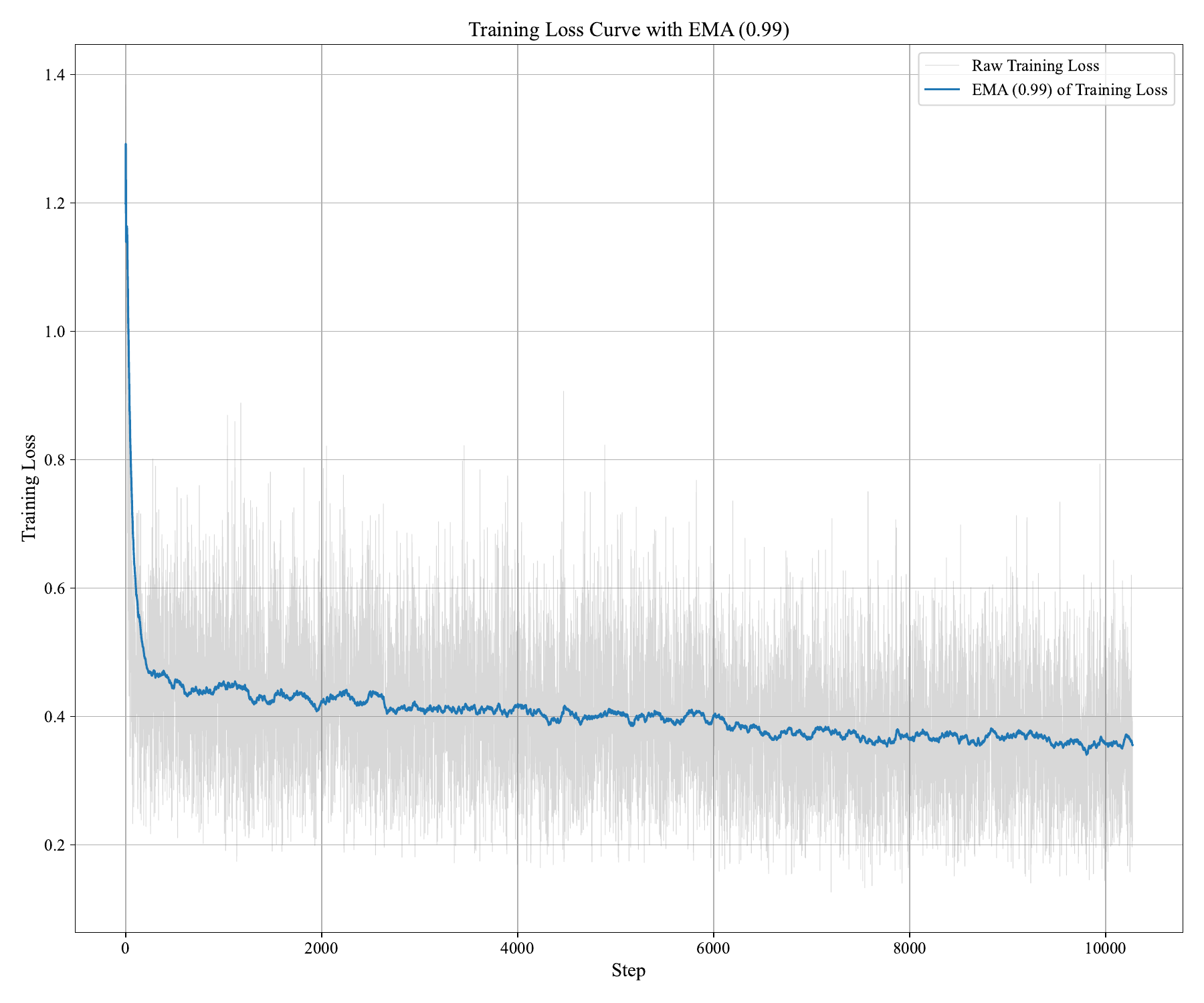}
  \vspace{1em}

  \includegraphics[height=0.28\textheight]{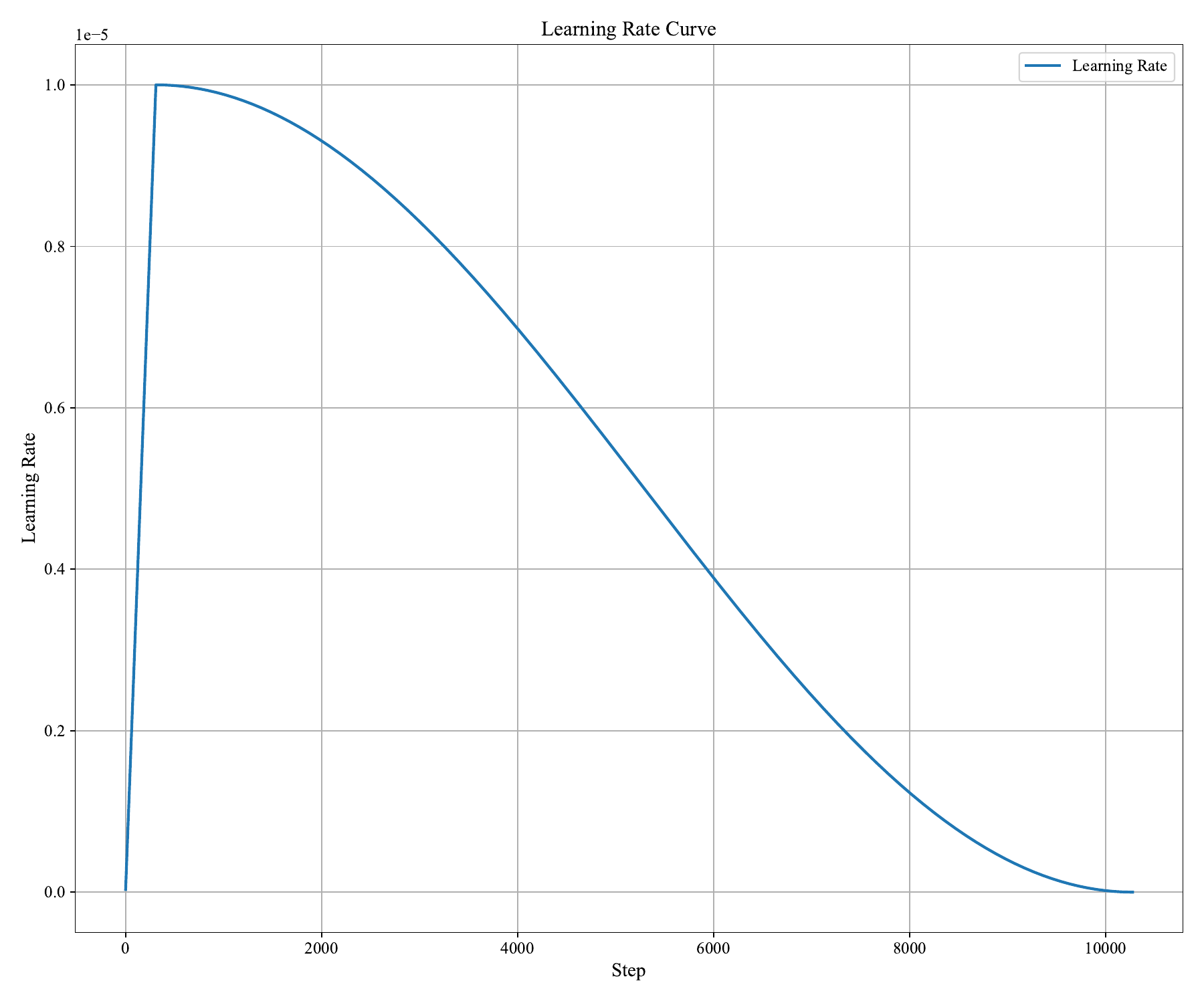}
  \vspace{1em}

  \includegraphics[height=0.28\textheight]{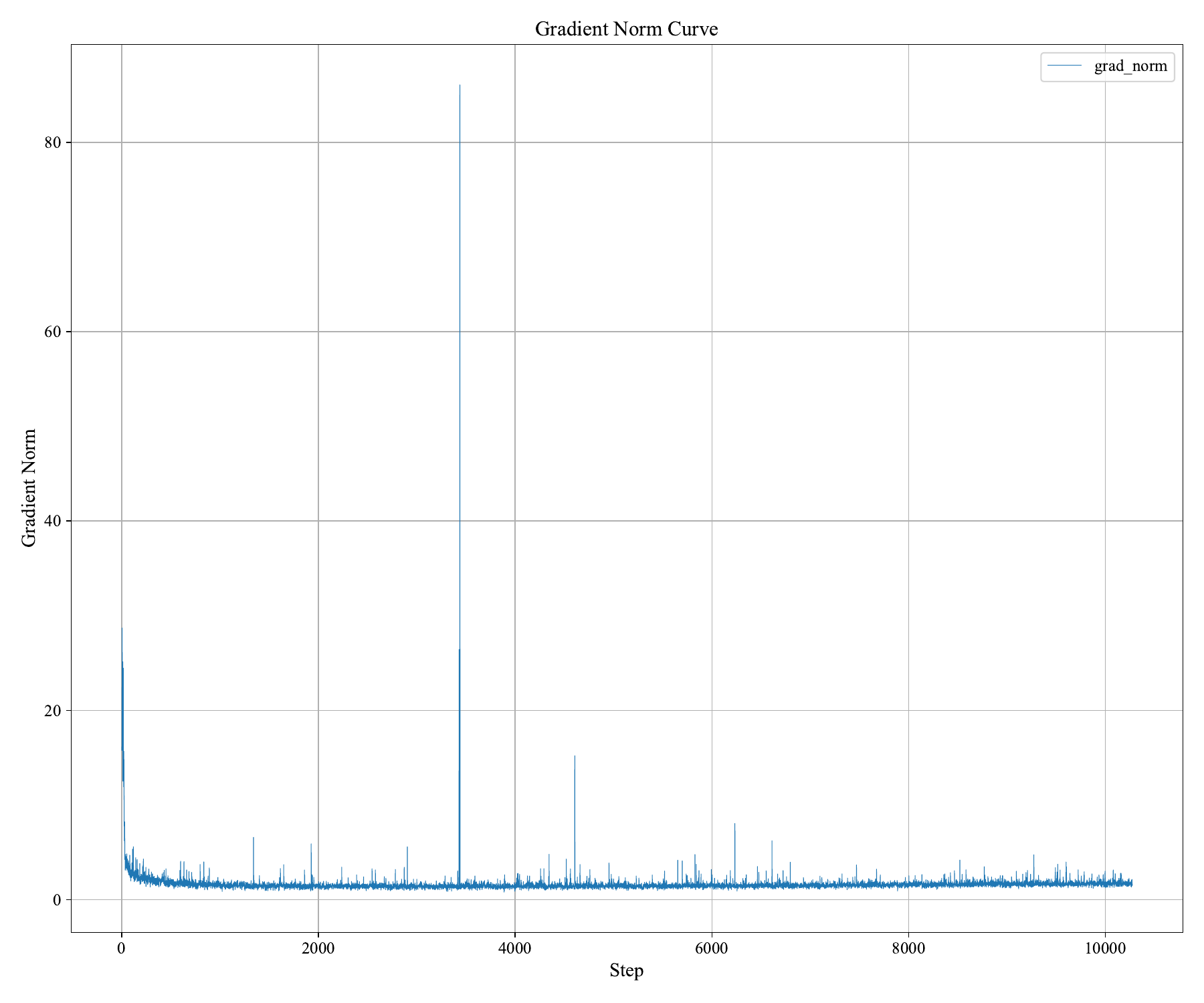}
  \caption{Training curves of ViCA-7B. From top to bottom: Training Loss with EMA, Learning Rate Schedule, and Gradient Norm.}
  \label{fig:training_curves_fullpage1}
\end{figure*}

\begin{figure*}[p]
  \centering
  \includegraphics[height=0.28\textheight]{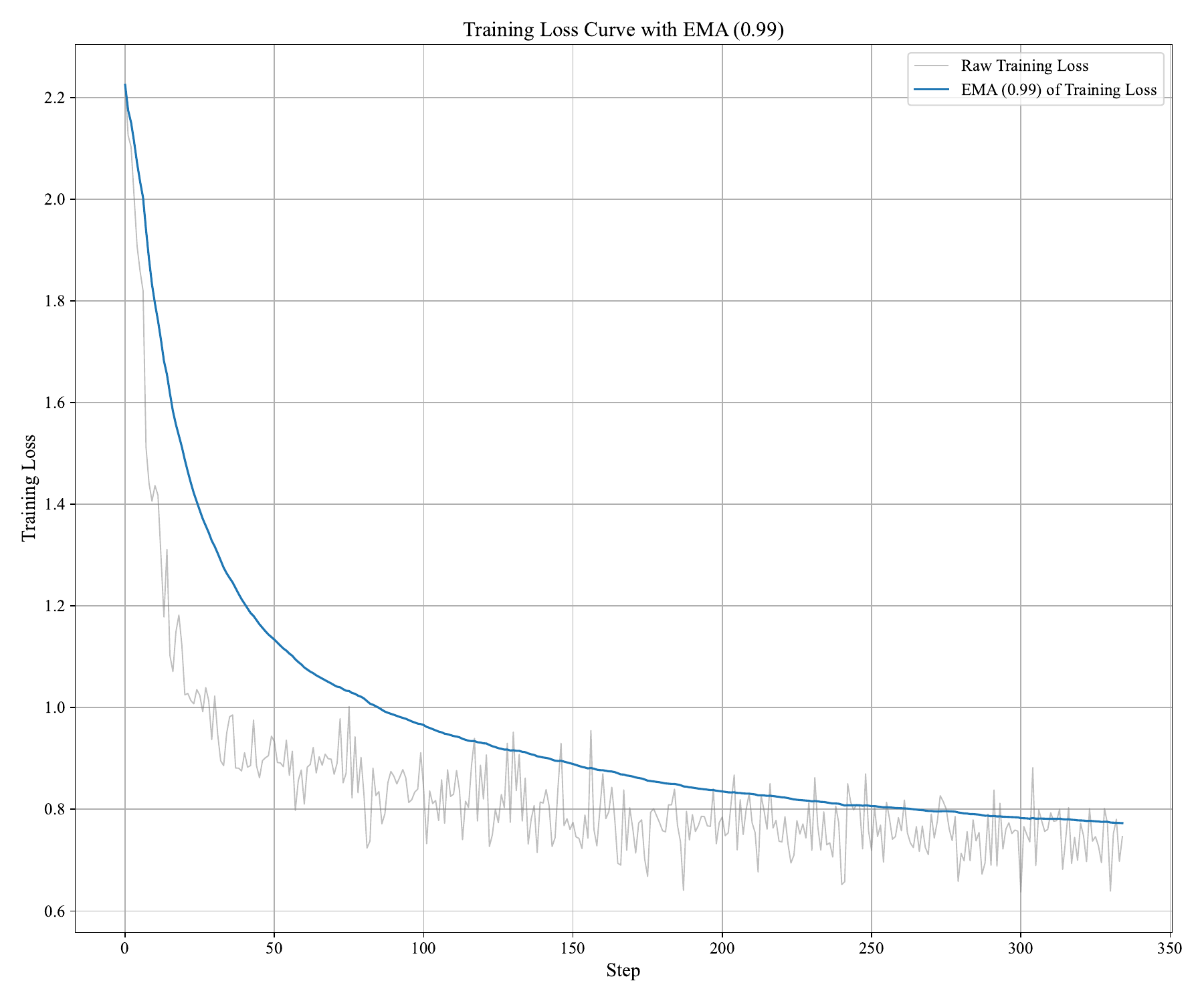}
  \vspace{1em}

  \includegraphics[height=0.28\textheight]{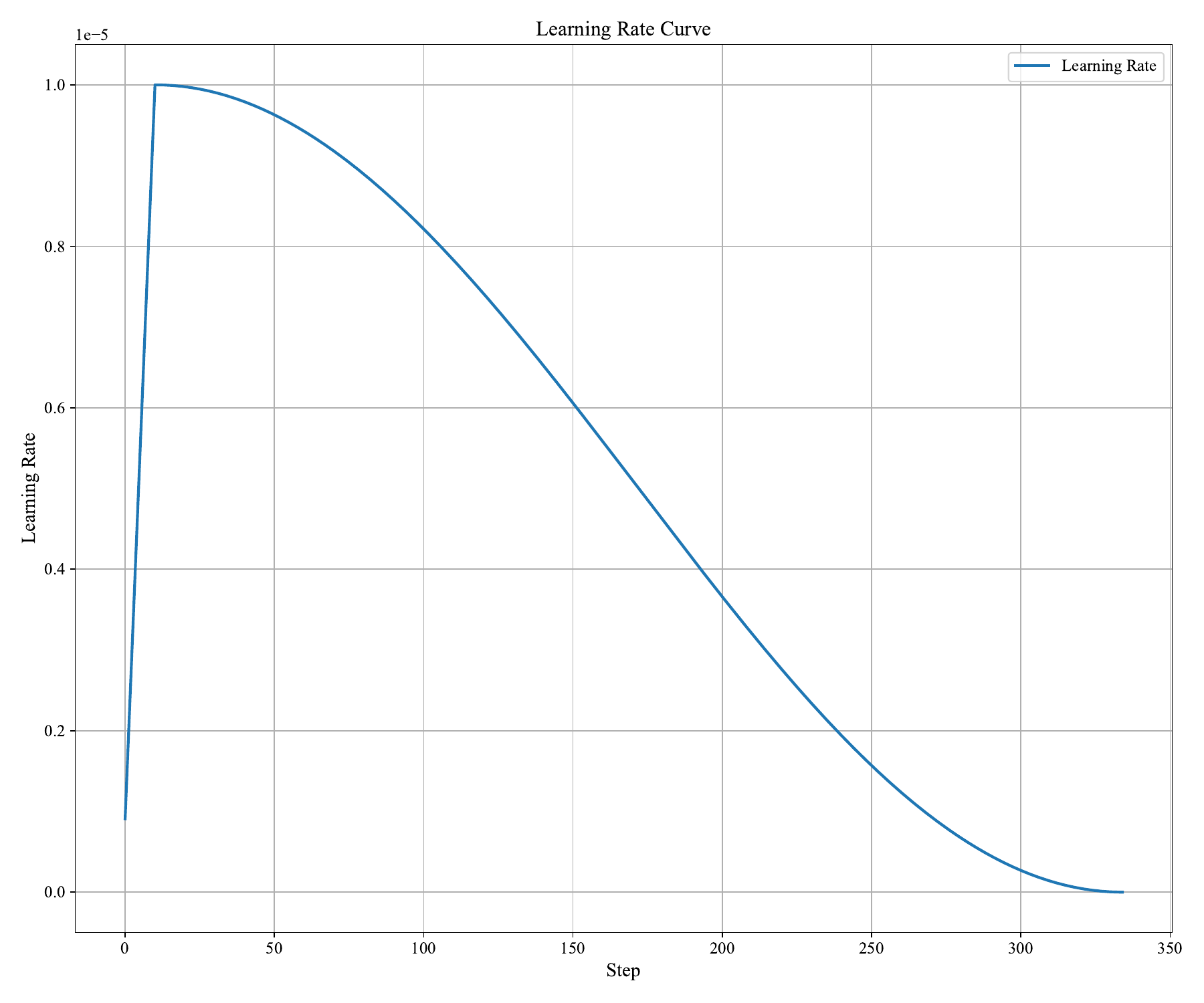}
  \vspace{1em}

  \includegraphics[height=0.28\textheight]{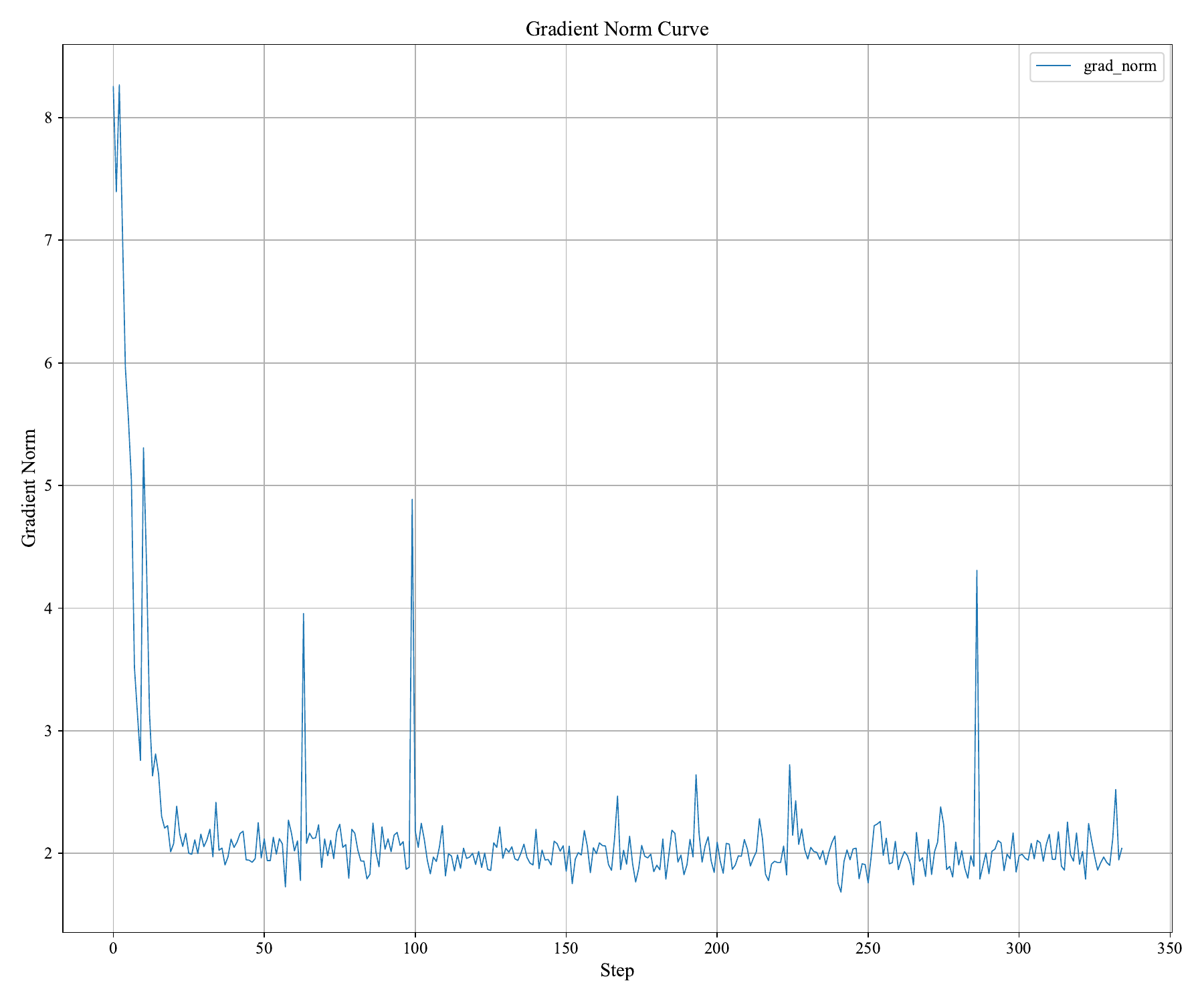}
  \caption{Training curves of ViCA-7B-Thinking. From top to bottom: Training Loss with EMA, Learning Rate Schedule, and Gradient Norm.}
  \label{fig:training_curves_fullpage2}
\end{figure*}

We publicly release the training logs(Figure~\ref{fig:training_curves_fullpage1}, \ref{fig:training_curves_fullpage2}) of both \textbf{ViCA-7B} and \textbf{ViCA-7B-Thinking} to support transparency and reproducibility. During the training of ViCA-7B, we observe a sharp decrease in training loss in the initial phase, followed by a stable and gradually declining trend. This pattern indicates that the model effectively learns from the ViCA-322K dataset and progressively refines its understanding over time.

For ViCA-7B-Thinking, the training loss remains relatively unstable even in the final stages of training. This is due to two factors: (1) the content of our \textbf{ViCA-Thinking-2.68K} dataset is significantly more complex and structurally different from that of ViCA-322K used in the previous fine-tuning stage; and (2) the small scale of the dataset introduces inherent variance, and there is still room for further loss reduction if the dataset were to be expanded. These trends reflect the unique challenges of fine-tuning for interpretable reasoning and suggest that additional data could further benefit learning in this setting.

\clearpage
\clearpage

\clearpage
\clearpage

\section{Analysis of Structured Output Behavior during ViCA-Thinking Fine-tuning}
\label{app:vica_thinking}

To further understand the learning dynamics of structured reasoning supervision, we conduct a series of controlled experiments on the ViCA-Thinking-2.68K dataset. Our goal is to investigate the following three questions:

\begin{itemize}
\item \textbf{Q1:} How much fine-tuning is required for the model to adopt the desired structured output format?
\item \textbf{Q2:} At different training stages, what proportion of outputs are syntactically valid and contain a usable final answer?
\item \textbf{Q3:} How does this structured fine-tuning impact the model's visuospatial reasoning performance, as measured by VSI-Bench?
\end{itemize}

To answer these questions, we train \textbf{ViCA-7B-Thinking} on ViCA-Thinking-2.68K and save checkpoints every 5\% of training progress. Each checkpoint is evaluated on the full VSI-Bench test set (5,130 examples).

\paragraph{Structured Output Format.}
The desired output format consists of three clearly segmented sections:

\begin{quote}
<|im\_start|>think
...\\
<|im\_start|>response
...\\
<|im\_start|>final
\end{quote}

We first examine whether the model produces outputs in this structured format. When using the original ViCA-7B (i.e., with 0\% ViCA-Thinking data), the model correctly predicts numerical or multiple-choice answers but does not generate any intermediate reasoning or formatted segmentation. After fine-tuning on 5\% of the data, we observe partial adoption of the desired structure: some outputs begin with a \texttt{<|im\_start|>think} segment, while others resemble unstructured predictions.

As shown in Table~\ref{tab:structured-output-progress}, the adoption of the structured format stabilizes after 15\% of training, with over 85\% of outputs across all VSI-Bench examples conforming to our defined format. This indicates that even limited exposure to structured reasoning examples is sufficient to induce the desired output behavior.

\paragraph{Syntactic Validity and Final Answer Presence.}
Not all outputs are valid even when the overall format is correct. Some responses are incomplete (e.g., missing a final answer section), contain malformed LaTeX (e.g., incorrect \verb|\boxed{}| syntax), or omit the final answer altogether. We define a response as \textit{valid} if it includes a properly formatted \texttt{final} section with a clearly extractable answer in one of the expected forms.

Table~\ref{tab:structured-output-progress} shows the proportion of valid responses across checkpoints. We observe steady improvements in syntactic and semantic correctness, plateauing after 15\%-20\% of training.

\paragraph{Answer Extraction and Evaluation.}
To evaluate model performance quantitatively, we extract the content within the following LaTeX-styled segment:

\begin{quote}

\begin{verbatim}
\[
\boxed{\text{Your final answer}}
\]
\end{verbatim}

\end{quote}

While most final answers are numerical or categorical, some contain vague language, ranges, or textual justification. To ensure evaluation consistency, we employ \texttt{Qwen2.5-32B-Instruct} as a post-processing model to canonicalize final answers into a numerical or multiple-choice format suitable for VSI-Bench scoring. This conversion process is fast and efficient: using a single NVIDIA H200, we process all 5,130 valid examples from a given checkpoint in under 13 minutes.

\paragraph{Performance Results.}
The results, presented in Table~\ref{tab:thinking-performance-scaling}, show a sharp drop in VSI-Bench accuracy after structured fine-tuning. Beyond 10\% of training on ViCA-Thinking-2.68K, the average score across all tasks falls below 40 and remains consistently low at higher percentages. These findings align with our main paper observations (see Section~\ref{sec:qualitative_vica_thinking}): although structured supervision enhances reasoning transparency, it introduces a measurable tradeoff in token-level task performance under current modeling frameworks.

\begin{table*}[t]
\small
\centering
\setlength{\tabcolsep}{4pt}
\begin{tabular}{c|r|r|r}
\toprule
\textbf{Training \%} & \textbf{Valid Outputs} & \textbf{Direct Answer} & \textbf{Via Thoughts} \\
\midrule
5   & 5036 & 4717 & 319 \\
10  & 4434 & 515  & 3919 \\
15  & 4432 & 0    & 4432 \\
20  & 4241 & 0    & 4241 \\
25  & 4438 & 0    & 4438 \\
30  & 3668 & 0    & 3668 \\
35  & 4313 & 0    & 4313 \\
40  & 4424 & 0    & 4424 \\
45  & 4338 & 0    & 4338 \\
50  & 4555 & 0    & 4555 \\
55  & 4524 & 0    & 4524 \\
60  & 4518 & 0    & 4518 \\
65  & 4559 & 0    & 4559 \\
70  & 4357 & 0    & 4357 \\
75  & 3826 & 0    & 3826 \\
80  & 4281 & 0    & 4281 \\
85  & 4414 & 0    & 4414 \\
90  & 4493 & 0    & 4493 \\
95  & 4496 & 0    & 4496 \\
100 & 4491 & 0    & 4491 \\
\bottomrule
\end{tabular}

\caption{
Structured output adoption across training progress. Each row shows the number of VSI-Bench test examples producing a valid output—either a direct answer or a structured output with explicit \texttt{<|im\_start|>think} and \texttt{<|im\_start|>response} segments—at a given training percentage.
}
\label{tab:structured-output-progress}
\end{table*}

\begin{table*}[t]
\small
\centering
\setlength{\tabcolsep}{3pt}
\renewcommand{\arraystretch}{1.15}
\resizebox{\textwidth}{!}{
\begin{tabular}{l|cccccccccccccccccccc}
\toprule
\textbf{Training \%} & \textbf{5} & \textbf{10} & \textbf{15} & \textbf{20} & \textbf{25} & \textbf{30} & \textbf{35} & \textbf{40} & \textbf{45} & \textbf{50} & \textbf{55} & \textbf{60} & \textbf{65} & \textbf{70} & \textbf{75} & \textbf{80} & \textbf{85} & \textbf{90} & \textbf{95} & \textbf{100} \\
\midrule
\textbf{Obj. Cnt}     & 65.35 & 46.85 & 41.24 & 34.48 & 28.97 & 40.42 & 36.61 & 40.23 & 39.85 & 37.77 & 37.28 & 34.12 & 35.10 & 40.24 & 37.51 & 36.51 & 34.36 & 37.85 & 36.70 & 37.32 \\
\textbf{Abs. Dist.}   & 52.16 & 24.48 & 26.47 & 26.18 & 26.24 & 27.53 & 28.00 & 27.98 & 28.25 & 25.60 & 26.16 & 24.96 & 27.21 & 28.31 & 28.89 & 29.51 & 28.20 & 29.63 & 28.48 & 29.10 \\
\textbf{Obj. Size}    & 76.00 & 55.25 & 41.18 & 37.92 & 42.27 & 40.05 & 42.90 & 46.51 & 42.82 & 41.28 & 48.80 & 44.25 & 47.49 & 46.17 & 45.55 & 45.65 & 48.94 & 46.34 & 46.12 & 46.94 \\
\textbf{Room Size}    & 63.19 & 21.14 & 25.58 & 30.88 & 22.29 & 26.80 & 36.53 & 32.60 & 31.16 & 29.23 & 37.52 & 29.93 & 35.37 & 32.33 & 35.75 & 35.56 & 31.15 & 32.54 & 33.05 & 31.68 \\
\textbf{Rel. Dist.}   & 58.31 & 45.66 & 40.28 & 47.17 & 44.77 & 43.27 & 47.47 & 46.98 & 43.01 & 43.31 & 45.35 & 48.16 & 45.05 & 48.01 & 41.84 & 48.77 & 49.79 & 50.61 & 46.91 & 50.20 \\
\textbf{Rel. Dir.}    & 37.95 & 34.73 & 37.13 & 36.41 & 37.91 & 34.63 & 39.06 & 36.74 & 36.72 & 34.89 & 36.43 & 38.61 & 39.18 & 32.68 & 36.55 & 37.83 & 39.84 & 40.21 & 37.86 & 35.95 \\
\textbf{Route Plan}   & 30.00 & 32.32 & 25.15 & 30.56 & 29.10 & 28.57 & 26.88 & 29.53 & 25.00 & 23.03 & 28.46 & 25.33 & 24.83 & 21.85 & 23.53 & 26.35 & 25.79 & 28.67 & 30.92 & 21.95 \\
\textbf{Appr. Order}  & 63.53 & 43.67 & 44.22 & 57.59 & 47.52 & 44.94 & 44.25 & 46.69 & 48.76 & 48.39 & 53.89 & 51.90 & 56.77 & 53.71 & 53.56 & 53.28 & 51.57 & 51.16 & 54.81 & 54.88 \\
\midrule
\textbf{Average}      & 55.81 & 38.01 & 35.16 & 37.65 & 34.88 & 35.78 & 37.71 & 38.41 & 36.95 & 35.44 & 39.23 & 37.16 & 38.87 & 37.91 & 37.90 & 39.18 & 38.70 & 39.63 & 39.36 & 38.50 \\
\bottomrule
\end{tabular}
}
\caption{
VSI-Bench performance of ViCA-7B-Thinking at different training stages (5\% to 100\% of ViCA-Thinking-2.68K). Each row reports the model's accuracy on a specific task, and the final row shows the average across all eight tasks. While structured supervision enables explainable reasoning (see Section~5.5), it introduces a notable decline in raw accuracy, particularly in early training stages.
}
\label{tab:thinking-performance-scaling}
\end{table*}

\clearpage
\onecolumn

\begin{center}
\Large\textbf{Prompt Used for Converting Outputs from ViCA-7B-Thinking}
\end{center}

\vspace{1em}

\begin{quote}

\label{tab:appendix_prompt}
\begin{Verbatim}[breaklines, fontsize=\small, frame=single]

You will be given a question and a response from a large language model (LLM) to that question.

Your task is to convert the LLM's response into a specific format I will describe.

There are two types of expected answers:

The correct answer should be a number, which can be either an integer or a decimal.
The correct answer should be a multiple-choice option, such as "A. xxx", "B. xxx", etc.
Here is how you should handle each case:

If the LLM's response is already a number, return it exactly as is, without any modifications.
Example: '2.12' → return '2.12'
If the response is already a multiple-choice option, return it exactly as is, without any modifications.
Example: 'B. front-right' → return 'B. front-right'
Example: 'C. ' → return 'C. '
If the response includes a number with units, strip off the units and return only the number.
Example: '17.5 square meters' → return '17.5'
If the response includes an estimated range, compute a reasonable estimate (e.g., the average) and return a single number without any unit.
Example: 'Approximately 10-11 square meters' → return '10.5'
Example: '50-60' → return '55'
If the response contains a textual explanation that indicates the answer cannot be determined or matched, return 'NULL'.
Example responses that should result in 'NULL':
'The refrigerator appears first, followed by the suitcase, and then the monitor. None of the options match this sequence.'
'The video does not show a chair, so the relative position cannot be determined.'
'The sequence of actions is Sink -> Washer -> Doorway, which does not match any of the options.'
'The question is flawed.'
In summary, your output must be one of the following three types only:

A number (e.g., '27.5')
A multiple-choice option (e.g., 'B. front-right')
'NULL'
Nothing else.

Now, here is the LLM's response for you to convert:

{repr(item['response'])}

Please output exactly one of the three allowed answer types mentioned above. Do not provide any explanation.
\end{Verbatim}
\end{quote}

\twocolumn

\label{appendix:qualitative}

\section{Evaluation: Effectiveness and Limitations of Structured Reasoning}

\subsection{Qualitative Evaluation: Improved Interpretability and Coherence}

To evaluate the effectiveness of structured reasoning supervision, we qualitatively analyze predictions made by our fine-tuned model, ViCA-7B-Thinking, on a subset of the VSI-Bench benchmark. In contrast to the original ViCA-7B—which is directly supervised to produce concise final answers—ViCA-7B-Thinking is instructed to articulate an explicit reasoning process in a dedicated Thoughts section before arriving at its final response.

\begin{figure*}[t]
\centering
\includegraphics[width=\linewidth]{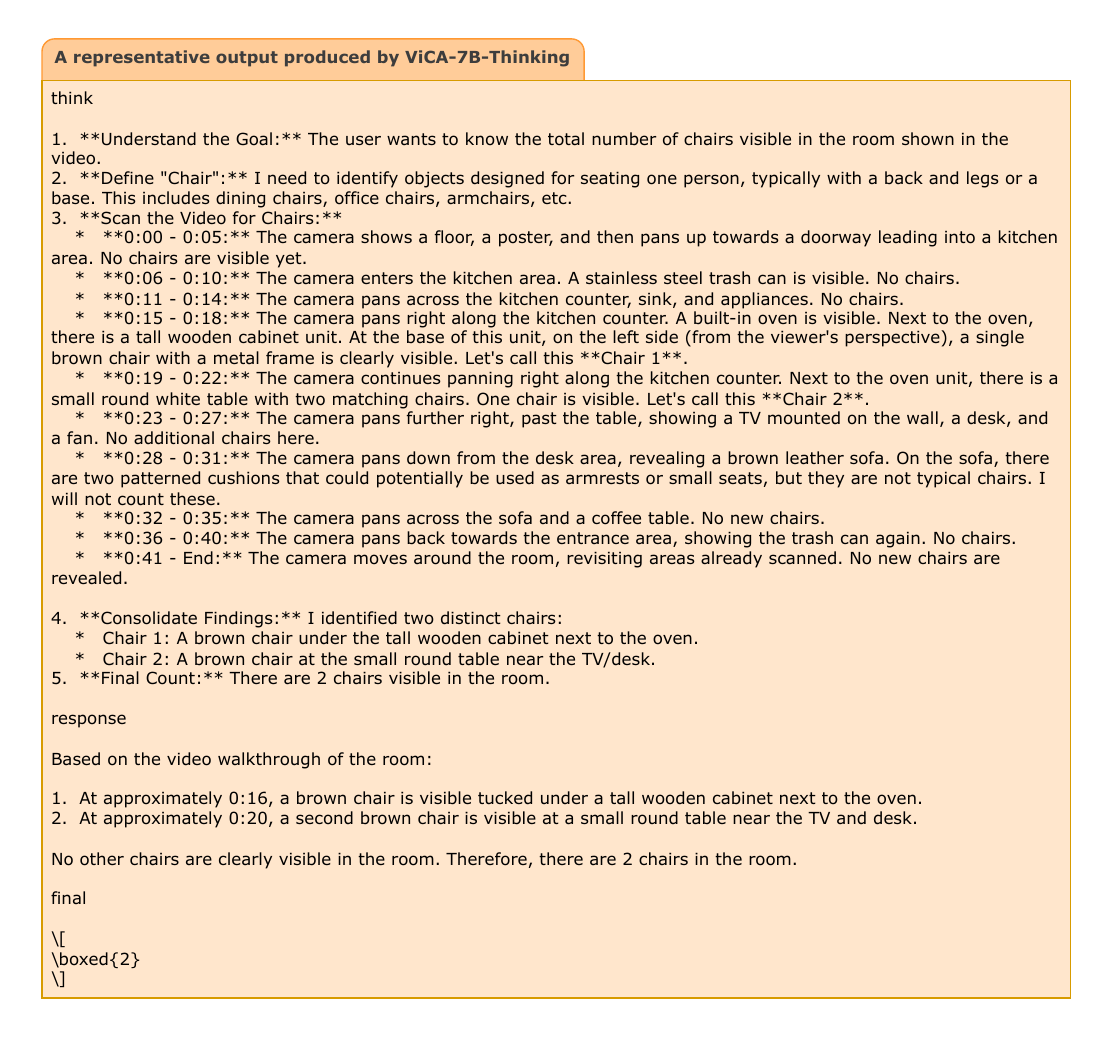}
\caption{A representative \textit{Thought-Response-Final Answer} example generated by ViCA-7B-Thinking. The model demonstrates its ability to externalize spatial reasoning in natural language.}
\label{fig:thought_example1}
\end{figure*}

\begin{figure*}[t]
\centering
\includegraphics[width=\linewidth]{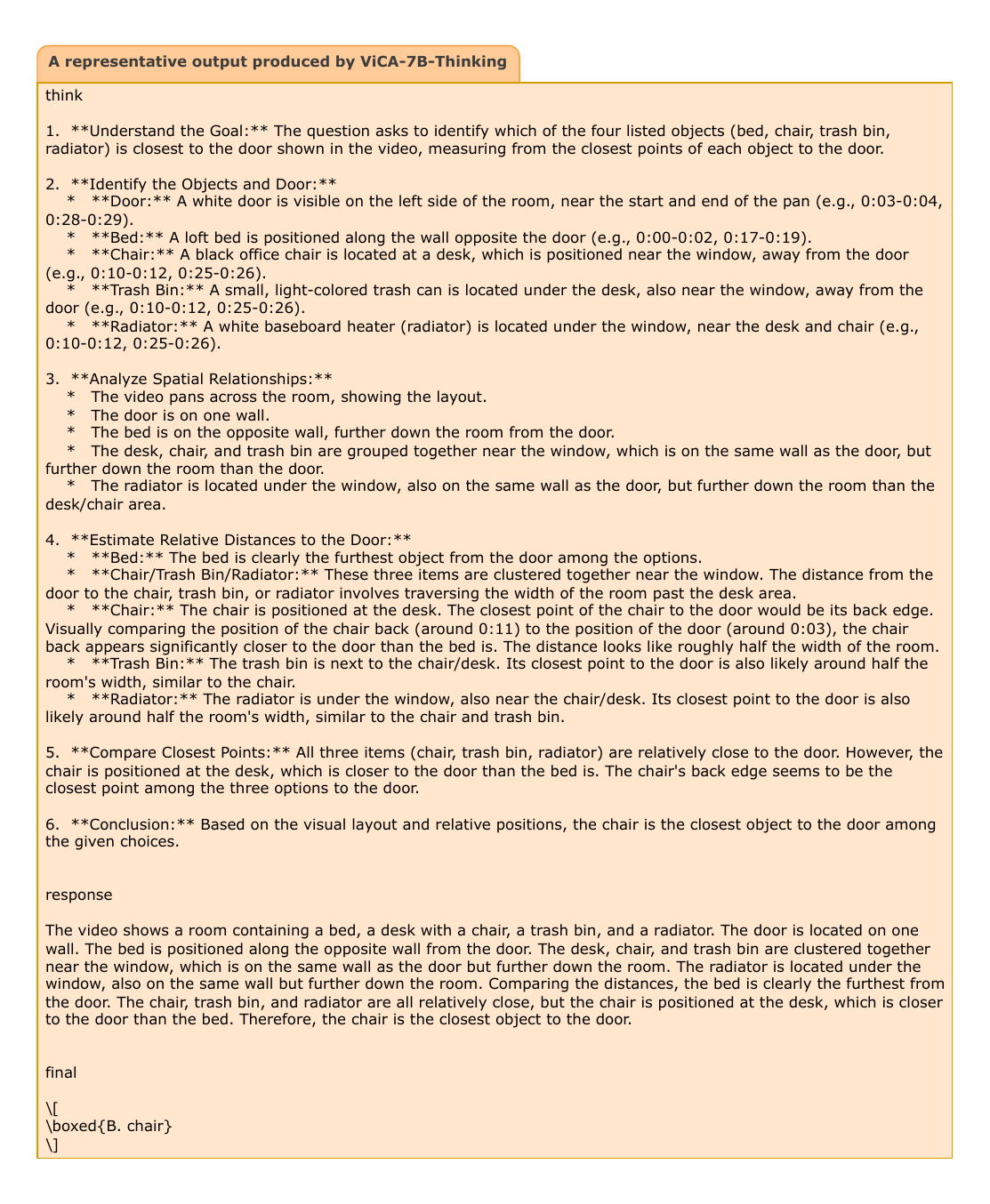}
\caption{A representative \textit{Thought-Response-Final Answer} example generated by ViCA-7B-Thinking. The model demonstrates its ability to externalize spatial reasoning in natural language.}
\label{fig:thought_example2}
\end{figure*}

\begin{figure*}[t]
\centering
\includegraphics[width=\linewidth]{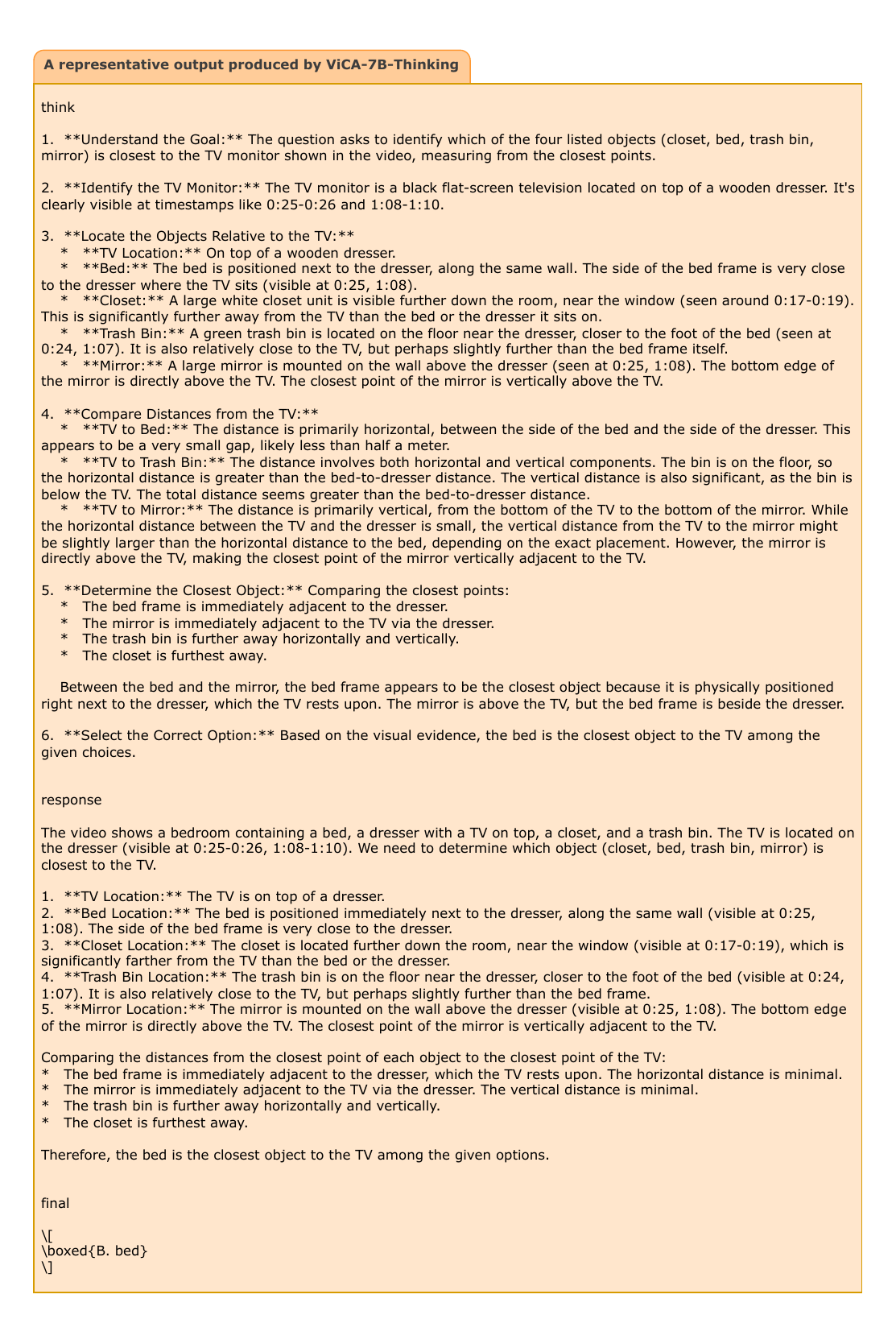}
\caption{A representative \textit{Thought-Response-Final Answer} example generated by ViCA-7B-Thinking. The model demonstrates its ability to externalize spatial reasoning in natural language.}
\label{fig:thought_example3}
\end{figure*}

\begin{figure*}[t]
\centering
\includegraphics[width=\linewidth]{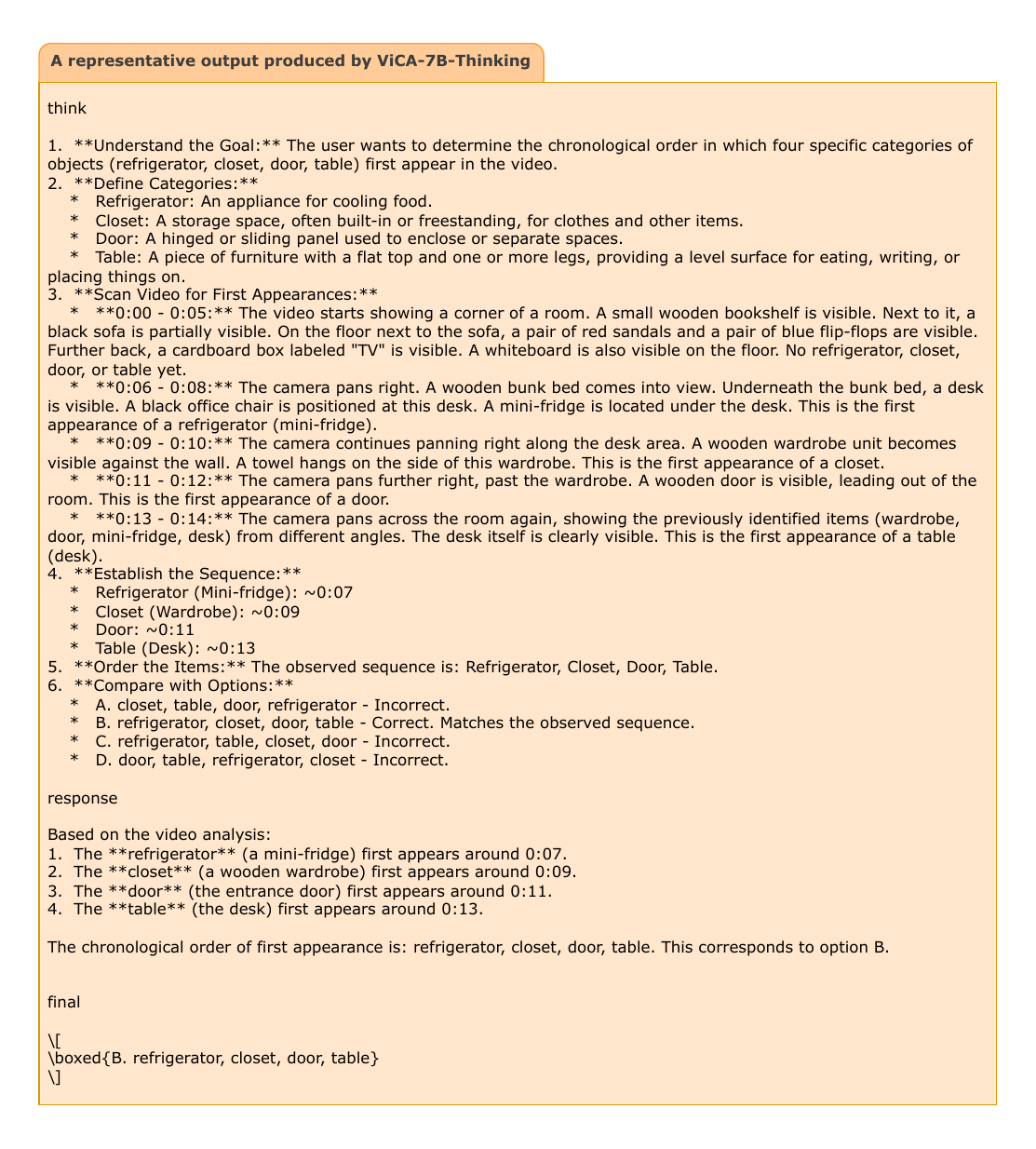}
\caption{A representative \textit{Thought-Response-Final Answer} example generated by ViCA-7B-Thinking. The model demonstrates its ability to externalize spatial reasoning in natural language.}
\label{fig:thought_example4}
\end{figure*}

Figure~\ref{fig:thought_example1}, \ref{fig:thought_example2}, \ref{fig:thought_example3} and \ref{fig:thought_example4} present representative cases that highlight the advantages of this approach. The model begins by identifying relevant visual cues from the input frames, proceeds to infer spatial relationships based on object positioning and scene geometry, and finally synthesizes these observations into a coherent conclusion. This progression mirrors a human-like chain of reasoning, where intermediate judgments are made explicit and traceable. Such outputs provide valuable interpretability: not only is the answer presented, but the why behind the answer becomes accessible to users and researchers.

Compared to the baseline ViCA-7B, which often relies on shortcut correlations learned from instruction tuning, ViCA-7B-Thinking demonstrates a higher degree of spatial awareness and contextual grounding in its verbalized reasoning. While this does not always result in higher factual accuracy (as discussed in Section~\ref{sec:qualitative_vica_thinking}), the structured format facilitates fine-grained error analysis, reveals latent model biases, and makes the reasoning process more amenable to human feedback and iterative correction.

\subsection{Quantitative Evaluation: Performance Tradeoffs and Insights}

To assess the quantitative impact of structured reasoning supervision, we evaluate ViCA-7B-Thinking on VSI-Bench and compare it with the original ViCA-7B. As summarized in Table~\ref{tab:vica-thinking-comparison}, ViCA-7B-Thinking obtains an average score of 38.5, which is 22.1 points lower than ViCA-7B (60.6). This decline is most pronounced in tasks requiring precise numerical predictions, such as Object Count (-31.5), Absolute Distance (-27.9), and Room Size (-43.4). Even in relatively less regression-heavy tasks such as Appearance Order and Route Planning, we observe noticeable drops (-13.9 and -12.5 respectively).

At first glance, this gap may suggest that structured reasoning negatively affects task performance. However, we argue that the observed degradation reveals a deeper structural limitation of current decoder-only architectures when adapting to multi-phase output formats. Specifically, requiring the model to generate both a coherent reasoning process (Thoughts) and a correct final answer in a single sequence may dilute the attention allocation and weaken the tight alignment previously learned between video input patterns and token-level outputs.

Nevertheless, we emphasize that this is not merely a loss in performance, but a constructive signal about the model's internal behavior. The structured format encourages the model to surface its latent reasoning processes, even if doing so disrupts its prior shortcut-driven optimization. As discussed in Section~\ref{sec:qualitative_vica_thinking}, ViCA-7B-Thinking produces substantially more interpretable outputs, where the connection between observations and conclusions is made explicit. In practice, such transparency is vital for applications where trust, auditability, and debugging are essential.

Moreover, our qualitative analysis indicates that ViCA-7B-Thinking tends to generalize better on under-specified or ambiguous inputs, and its errors are often more “reasonable” in the sense that they reflect plausible but imperfect reasoning chains. In contrast, ViCA-7B occasionally outputs correct answers without any visible justification—a behavior that may be beneficial for benchmarks but undesirable for safety-critical scenarios.

In summary, while ViCA-7B-Thinking underperforms its counterpart in raw accuracy, it offers a valuable shift toward explainable spatial cognition. We view this as an important step toward the broader goal of building multimodal models that can reason in human-like, interpretable ways—paving the way for future architectures capable of reconciling explicit reasoning with precise prediction.

\begin{table}[t]
  \small
  \centering
  \setlength{\tabcolsep}{4pt}
  \begin{tabular}{lcc}
    \toprule
    & \textbf{ViCA-7B} & \textbf{ViCA-7B-Thinking} \\
    \midrule
    \rowcolor{gray!20} \multicolumn{3}{l}{\textit{Numerical Answer}} \\
    Object Count       & \textbf{68.8} & 37.3 (-31.5) \\
    Absolute Distance  & \textbf{57.0} & 29.1 (-27.9) \\
    Object Size        & \textbf{79.2} & 46.9 (-32.3) \\
    Room Size          & \textbf{75.1} & 31.7 (-43.4) \\
    \midrule
    \rowcolor{gray!20} \multicolumn{3}{l}{\textit{Multiple-Choice Answer}} \\
    Relative Distance  & \textbf{58.5} & 50.2 (-8.3) \\
    Relative Direction & \textbf{42.6} & 36.0 (-6.6) \\
    Route Planning     & \textbf{34.5} & 22.0 (-12.5) \\
    Approach Order     & \textbf{68.8} & 54.9 (-13.9) \\
    \midrule
    \textbf{Average}   & \textbf{60.6} & 38.5 (-22.1) \\
    \bottomrule
  \end{tabular}
  \caption{
  Comparison between ViCA-7B and ViCA-7B-Thinking on spatial reasoning benchmarks. 
  The values in parentheses denote the performance difference from ViCA-7B.
  Bold highlights indicate the higher score between the two models for each metric.
  }
  \label{tab:vica-thinking-comparison}
\end{table}

\subsection{Summary and Takeaways}

Our investigation into structured reasoning supervision via ViCA-7B-Thinking reveals a nuanced tradeoff between interpretability and task-specific accuracy. While performance on standard evaluation metrics declines significantly—particularly on numerically grounded tasks—the model gains the ability to explicitly articulate its internal reasoning trajectory in a human-readable form. This enhancement is not merely cosmetic: it surfaces previously opaque decision-making processes and facilitates qualitative analysis, error attribution, and human-in-the-loop verification.

These findings underscore a key challenge in current decoder-only multimodal architectures: the difficulty of jointly optimizing for factual precision and reasoning coherence within a single generation stream. The observed performance degradation should therefore be interpreted not as a regression in modeling capability, but as evidence of structural friction in aligning generation objectives with cognitive transparency.

We believe that structured reasoning formats, despite their current limitations, represent a critical direction for the future of multimodal AI. To this end, we advocate for further research into hybrid generation schemes, modular decoding strategies, and task-specific reasoning heads that can disentangle explanation from prediction. Ultimately, our results suggest that interpretability and performance need not be at odds—if explicitly modeled, both may be improved in tandem.

\section{Prompt Template for the ViCA-Thinking-2.68K Dataset}

In this section, we provide the complete Prompt Template for the ViCA-Thinking-2.68K dataset.

\onecolumn

\begin{center}
\Large\textbf{Prompt Template for the ViCA-Thinking-2.68K Dataset}
\end{center}

\vspace{1em}

\begin{quote}

\label{app:appendix_prompt}
\begin{Verbatim}[breaklines, fontsize=\small, frame=single]
You are an expert AI assistant specializing in visuospatial reasoning. Please read the following video description carefully and generate 10 high-quality question-answer pairs that require fine-grained spatial understanding. Each question should be grounded in the content of the video, with emphasis on ... 

You will be given a question. Please answer it based on the provided video.

You are required to first go through a clear reasoning process and explicitly write out your thought process. While thinking, you should plan your approach and go through a series of logical deductions.

Below are some examples for your reference:

---

### Example 1

"Here's my thought process for determining the number of cabinets in the room based on the video:

1.  **Understand the Goal:** The request asks for the number of cabinets *in the room*. The video shows a person entering a room and panning around.

2.  **Define ""Cabinet"":** I need a working definition of ""cabinet"" for this context. Generally, a cabinet is a piece of furniture with shelves or drawers, enclosed by doors or open. This could include kitchen cabinets, bathroom cabinets, dressers, chests of drawers, storage units, etc. I need to be consistent in applying this definition.

3.  **Analyze the Video Frame by Frame (or segment by segment):**
    *   **0:00 - 0:05:** The camera looks down at the floor (black and white tiles) and feet wearing blue shoe covers. There's a comic book on the floor. The view then transitions to looking forward into a room with wood flooring. No cabinets visible yet, just the doorway area.
    *   **0:05 - 0:10:** The camera pans up and into the room. I see a bed, a window, a yellow hanging lamp, and part of a door.
    *   **0:10 - 0:14:** The camera pans left.
        *   A white **chest of drawers** (dresser) comes into view. It has multiple drawers (looks like 4 or 5) with wooden knobs. This fits the definition of a cabinet/storage unit. **Cabinet 1**.
        *   Next to it, a white **cube storage unit** (like an IKEA Kallax) is visible. It has multiple square compartments, some holding items (like a red suitcase, folded clothes/shoes). This is also a storage unit/cabinet. **Cabinet 2**.
    *   **0:14 - 0:21:** The camera continues panning left, then up towards the ceiling, showing the lamp again, then moves towards the window area. I briefly see the white chest of drawers and cube storage again from a different angle.
    *   **0:21 - 0:32:** The camera focuses on the window area. An air purifier (white, rectangular object) is visible, then a baby crib/bassinet. No new cabinets here.
    *   **0:32 - 0:38:** The camera pans across the room, showing the bed, the crib again.
    *   **0:38 - 0:41:** The camera focuses on the area next to the crib/bassinet, near the radiator. There's a tall, narrow, light **wooden bookshelf/shelving unit**. It has multiple shelves holding books and other items. This counts as a cabinet/storage unit. **Cabinet 3**.
    *   **0:41 - 0:51:** The camera pans back across the room, showing the bed, the white chest of drawers (Cabinet 1), and the cube storage (Cabinet 2) again. It passes a white chest/box on the floor near the door - maybe a toy box? It looks like a low storage chest. Let's count it. **Cabinet 4**. (It's the white box with wooden knobs on top seen clearly around 0:50).
    *   **0:51 - 0:55:** The camera pans into what seems like a different area or the other side of the room/closet area. It shows the cube storage (Cabinet 2) again, and next to it, a taller, light **wooden chest of drawers/dresser**. This one looks different from Cabinet 1 (it's wood, not white). It has multiple drawers. **Cabinet 5**.
    *   **0:55 - 1:05:** The camera pans around this area, showing a fan, a folding screen, jeans hanging on a hook, and the wooden dresser (Cabinet 5) again.
    *   **1:05 - 1:18:** The camera looks towards a slanted ceiling with a skylight/roof window. It shows the top of the cube storage (Cabinet 2) and the wooden dresser (Cabinet 5) again. Some hangers are visible near the window.
    *   **1:18 - 1:27:** The camera pans back down, showing the white chest of drawers (Cabinet 1) and the door again. The low white chest/box (Cabinet 4) is visible again near the door.
    *   **1:27 - 1:38:** The camera moves back out of the room, showing the doorway and the patterned rug/mat outside the room, then back towards the entrance area with the black and white tiles.
    *   **1:38 - 1:41:** The camera focuses back on the starting point, looking down at the comic book and feet.

4.  **Consolidate and Count:**
    *   Cabinet 1: White chest of drawers/dresser (seen multiple times, e.g., 0:11, 0:45, 1:20).
    *   Cabinet 2: White cube storage unit (seen multiple times, e.g., 0:12, 0:46, 0:52, 1:11).
    *   Cabinet 3: Tall, narrow wooden bookshelf/shelving unit (seen near the radiator/crib, e.g., 0:39).
    *   Cabinet 4: Low white storage chest/box (seen near the door, e.g., 0:50, 1:22).
    *   Cabinet 5: Taller wooden chest of drawers/dresser (seen next to cube storage, e.g., 0:53, 1:06).

5.  **Final Check:** Review the list against the video segments. Did I miss any? Did I double-count? The identified items seem distinct and fit the definition of cabinets/storage furniture. The count is 5.

6.  **Formulate the Answer:** State the final count clearly."

---

### Example 2

1.  **Analyze the Request:** The user wants to know which object (sink, bathtub, bed, or toilet) is nearest to the ""cabinet"" shown in the video. The answer format requires the letter followed by the full option.

2.  **Identify the ""Cabinet"":** The video starts focused on a white, built-in structure that looks like a cabinet or perhaps access to storage/utilities under the eaves of the roof (common in loft conversions). It has double doors. Let's assume this is the ""cabinet"" referred to.

3.  **Track the Camera Movement:**
    *   0:00-0:04: Close-up of the cabinet door opening, looking *into* the bathroom. The camera is positioned *outside* the bathroom, near the cabinet.
    *   0:05-0:07: Camera pans left, moving *into* the bathroom. The sink comes into view on the left wall.
    *   0:07-0:09: Camera continues panning left, showing the shower area next to the sink.
    *   0:09-0:14: Camera pans further left/down, showing the toilet and the area behind it (shower controls, niche).
    *   0:15-0:18: Camera pans back towards the sink and the entrance door. The sink is clearly visible, along with a heated towel rail next to the door.
    *   0:19-0:25: Camera continues panning, focusing on the shower area again.
    *   0:26-0:39: Camera moves to the far end of the bathroom, showing the toilet and the freestanding bathtub next to it.
    *   0:40-0:44: Camera looks up towards the skylight and back towards the entrance door.
    *   0:45-0:52: Camera focuses on the bathtub again.
    *   0:53-0:57: Camera shows a small window near the bathtub.
    *   0:58-1:04: Camera moves back towards the entrance, showing the sink, towel rail, and the view into the bedroom through the open door. The cabinet is visible again near the entrance.
    *   1:05-1:18: More views of the bathtub, toilet, and panning back towards the entrance.

4.  **Assess Proximity:**
    *   **Cabinet Location:** The cabinet is located just outside the main bathroom area, near the entrance door, under a sloped ceiling.
    *   **Sink:** The sink is immediately to the left as you enter the bathroom, making it very close to the entrance and thus relatively close to the cabinet just outside the entrance.
    *   **Toilet:** The toilet is further into the bathroom, past the shower area, at the far end relative to the entrance.
    *   **Bathtub:** The bathtub is at the very far end of the bathroom, next to the toilet, furthest from the entrance and the cabinet.
    *   **Bed:** The bed is in the adjacent room (bedroom), seen through the open doorway. While the doorway is near the cabinet, the bed itself is further inside the bedroom.

5.  **Compare Distances:**
    *   The cabinet is right by the entrance.
    *   The sink is just inside the entrance.
    *   The toilet and bathtub are at the far end.
    *   The bed is in another room.

6.  **Conclusion:** The sink is the closest object *within the bathroom* to the location of the cabinet (which is just outside/at the entrance). Comparing the sink to the bed requires considering the threshold. The cabinet is right next to the door frame leading into the bedroom. The sink is just inside the bathroom door frame. Visually, the sink appears slightly closer to the cabinet's location than the bed does, especially considering the bed is set back within the bedroom. However, the cabinet *is* part of the structure that separates the bathroom entrance area from the bedroom. The sink is *inside* the bathroom proper, while the bed is *inside* the bedroom proper. The cabinet sits *between* these two areas, very near the threshold.

7.  **Re-evaluate Cabinet Definition:** Is the ""cabinet"" the structure seen initially, or could it refer to something else? No other cabinet is prominently featured. The initial structure seems the most likely candidate. It appears to be built into the wall/eaves space *between* the bedroom and the bathroom entrance area.

8.  **Refine Proximity Comparison:**
    *   The cabinet is structurally part of the wall separating the bedroom/hallway from the bathroom entrance.
    *   The bed is in the bedroom, adjacent to this wall/cabinet.
    *   The sink is in the bathroom, adjacent to the opposite side of the doorway near the cabinet.
    *   The toilet and bathtub are much further away.

    Looking closely at 0:00-0:04 and 1:02-1:04, the cabinet seems to be *in* the bedroom space, right next to the bathroom door. The camera starts outside the bathroom, looking *through* the cabinet opening into the bathroom. Then it enters the bathroom. When looking back out (1:02-1:04), the sink is to the left of the door, the towel rail is between the sink and the door, and the bed is visible through the door. The cabinet is part of the wall structure the door is set into. Therefore, the objects closest to the *physical structure* of the cabinet are the bed (on the bedroom side) and the sink/towel rail (on the bathroom side, near the door).

    Let's measure visually from the *center* of the cabinet doors:
    *   Distance to sink: Requires passing through the doorway and moving slightly left.
    *   Distance to bed: Seems very close, just on the other side of the wall/cabinet structure.
    *   Distance to toilet/bathtub: Clearly much further.

    The framing around the cabinet opening (0:02-0:03) suggests it might be built into the wall separating the bedroom and bathroom. The perspective at 1:03 shows the bathroom door opening *into* the bedroom. The cabinet is on the wall to the right of the bathroom door (when viewed from the bedroom). The sink is inside the bathroom, to the left of the door frame. The bed is in the bedroom, likely against the wall opposite the bathroom door or the wall adjacent to the cabinet.

    Consider the path: From the cabinet -> doorway -> sink. From the cabinet -> turn around -> bed. Given the cabinet is built into the wall *between* the rooms, both the bed and the sink are nearby, but likely on opposite sides of the wall/doorway.

    Let's re-examine the very beginning. The camera seems positioned in the bedroom, looking towards the bathroom *through* the open cabinet doors, which are set into the wall. The bathroom is reflected. Then the camera pulls back (0:01-0:03) revealing the cabinet is built into the wall of the bedroom, facing the bathroom door. The camera then enters the bathroom (0:04 onwards).

    Okay, new interpretation:
    *   The camera starts *in the bedroom*.
    *   The cabinet is built into the bedroom wall, adjacent to the bathroom door.
    *   The reflection shows the bathroom.
    *   The camera moves towards and enters the bathroom.
    *   The sink is immediately inside the bathroom.
    *   The bed is in the bedroom where the camera started, near the cabinet.

    Comparing distance from the physical cabinet structure:
    *   Bed: In the same room, likely very close.
    *   Sink: In the adjacent room, just through the doorway.
    *   Toilet/Bathtub: Far away in the bathroom.

    Therefore, the bed is likely the nearest object to the physical cabinet structure.

9.  **Final Check:** The video shows the cabinet is in the bedroom space, right next to the bathroom door frame (visible from 0:01-0:03 and 1:02-1:04). The bed is also in the bedroom (seen clearly through the door at 1:02-1:04). The sink is just inside the bathroom. Given the cabinet and bed are in the same room, and the cabinet is adjacent to the door, the bed is most likely the closest of the options listed.

10. **Format the Answer:** The letter is C, and the full option is ""bed""."

---

### Example 3

"The user wants to identify the first appearance time of four specific categories (refrigerator, chair, tv_monitor, oven) in the video and determine their sequence.

1.  **Refrigerator:** Appears early, visible through the doorway around 0:01-0:02, more clearly seen as the camera moves forward around 0:06-0:07. It's the large stainless steel appliance on the left around 03:08. Let's pinpoint the first *clear* view. It's partially visible from the start through the doorway, but the first clear, unobstructed view seems to be around 03:08 as the camera pans left. However, it is visible much earlier as part of the kitchen setup, seen beyond the dining table. Let's re-evaluate. Ah, wait, the *stainless steel* one is seen clearly at 03:08, but there's a *different* fridge/freezer integrated into the cabinetry earlier? No, the large stainless steel one is the main one. Looking again at the start, through the doorway, you see the kitchen area. A tall, light-coloured appliance (possibly integrated fridge/freezer) is visible against the back wall from 0:00. Let's call the first visible instance of *any* refrigerator. Yes, a tall unit, likely a refrigerator/freezer, is visible in the kitchen background from the very start (0:00).

2.  **Chair:** Chairs are visible almost immediately. The dining chairs around the table are clearly seen starting around 0:05-0:06. There's also an armchair near the window visible from 0:08-0:09. A computer chair appears later around 1:31. The dining chairs are the first type seen.

3.  **TV Monitor:** An iMac computer monitor is visible on a desk around 1:31-1:32. Is there another TV earlier? Yes, there appears to be a screen, maybe a TV or large monitor, mounted on the wall or sitting high up near the red LED strip, visible from around 0:07-0:08. Let's confirm. Yes, at 0:07, a black rectangular screen is visible above the kitchen counter area, framed by red light. This seems to be the first instance. The iMac appears much later.

4.  **Oven:** An oven (part of a range cooker, looks like an AGA or similar) is visible in the kitchen area. It's black and noticeable from around 0:07 when the kitchen area comes into view. It's more clearly seen later, around 03:03 and especially 03:40. The first appearance is around 0:07.

**Sequence Determination:**

*   **Refrigerator:** Visible in the background from 0:00.
*   **Chair:** Dining chairs clearly visible from ~0:06.
*   **TV Monitor:** Screen near red light visible from ~0:07.
*   **Oven:** Range cooker visible from ~0:07.

Okay, refining the times:
*   **Refrigerator:** 0:00 (visible through doorway in kitchen)
*   **Chair:** 0:06 (dining chairs)
*   **TV Monitor:** 0:07 (screen near red light)
*   **Oven:** 0:07 (black range cooker)

The sequence is: Refrigerator, Chair, TV Monitor/Oven (appear almost simultaneously).

Let's check the options against this sequence: Refrigerator, Chair, TV Monitor, Oven.

*   A. chair, tv_monitor, refrigerator, oven - Incorrect (Refrigerator is first)
*   B. refrigerator, chair, tv_monitor, oven - Matches the observed sequence.
*   C. tv_monitor, oven, chair, refrigerator - Incorrect (Refrigerator is first)
*   D. oven, refrigerator, tv_monitor, chair - Incorrect (Refrigerator is first)

Double-checking the very first frames for any chairs. The view starts looking through a doorway. No chairs are clearly visible *before* the dining chairs appear around 0:06. The refrigerator is visible *through* the doorway in the kitchen area right from 0:00. The TV Monitor and Oven become visible around 0:07 as the camera view encompasses more of the kitchen/dining area.

Therefore, the sequence is Refrigerator (0:00), Chair (0:06), TV Monitor (0:07), Oven (0:07). Option B aligns with this."

---

### Example 4

"Here's my thought process for estimating the sofa's length:

1.  **Understand the Goal:** The request asks for the length of the *largest side* of the sofa in centimeters. This means I need to identify the longest dimension of the sofa visible in the video and estimate its measurement.

2.  **Analyze the Video:**
    *   The video starts with a close-up of the floor, showing a movie poster printout (""Blackmail"") and someone's feet wearing blue shoe covers. This doesn't help with the sofa measurement yet.
    *   The camera pans up and moves through a doorway into a living room (0:05 - 0:07).
    *   The camera pans around the living room, showing a sofa against one wall (0:08 - 0:21). This is the primary view of the sofa.
    *   The camera continues panning, showing the window, another chair, the TV area, furniture, and eventually returns to the doorway area.

3.  **Identify the Sofa and its Largest Side:** The sofa is the long, dark grey piece of furniture pushed against the light grey wall. Its largest side is clearly its length, running parallel to the wall.

4.  **Find Reference Objects:** To estimate the sofa's length, I need objects with known or easily estimated dimensions within the same frame or perspective. Potential reference objects:
    *   **Floorboards:** The floor appears to be laminate or wood plank flooring. Standard plank widths are often around 12-19 cm (5-7.5 inches).
    *   **Doorway:** Standard interior doorways are often around 80 cm wide. However, the doorway isn't directly next to the sofa in a way that makes comparison easy.
    *   **Light Switch:** Standard light switch plates are roughly 8 cm wide and 12 cm tall.
    *   **Electrical Outlet:** Similar dimensions to the light switch plate.
    *   **Window:** Windows vary greatly.
    *   **TV:** TV screen sizes are measured diagonally, and widths vary.
    *   **Armchair:** The armchair (looks like an IKEA Poäng or similar) has a typical width of around 68 cm.
    *   **Coffee Table:** The white coffee table looks like a standard small IKEA Lack table, which is 55x55 cm.
    *   **Poster (""Fenêtre sur Cour""):** This is a standard movie poster size, maybe A1 (59.4 x 84.1 cm) or similar, but it's hard to tell for sure, and perspective makes it difficult to use accurately.
    *   **Person (briefly seen in doorway):** People vary too much in height/width.
    *   **Ceiling Height:** Standard ceiling heights are often around 240-270 cm, but this is hard to relate directly to the sofa length.

5.  **Select the Best References and Estimate:**
    *   **Floorboards:** Let's count the floorboards along the length of the sofa (around 0:18-0:20). It looks like the sofa spans approximately 10-12 full floorboard widths. If we assume an average width of 15 cm per board: 10 * 15 cm = 150 cm; 12 * 15 cm = 180 cm. If we assume a wider board, say 19 cm: 10 * 19 cm = 190 cm; 12 * 19 cm = 228 cm. This gives a range, maybe centering around 180-200 cm.
    *   **Coffee Table:** The coffee table (likely 55 cm wide) is placed on a rug. Comparing the sofa length to the table width is tricky due to perspective, but the sofa looks significantly longer, perhaps 3 to 4 times the table width. 3 * 55 cm = 165 cm; 4 * 55 cm = 220 cm. This aligns roughly with the floorboard estimate.
    *   **Armchair:** The armchair (est. 68 cm wide) is next to the coffee table. Again, comparing directly is hard, but the sofa appears roughly 2.5 to 3 times the width of the armchair. 2.5 * 68 cm = 170 cm; 3 * 68 cm = 204 cm. Consistent range.
    *   **Sofa Type:** It looks like a standard 3-seater sofa, possibly a sofa bed. These typically range from 180 cm to 240 cm in length.

6.  **Refine the Estimate:** The different reference points (floorboards, coffee table, armchair, typical sofa size) seem to converge on a range roughly between 170 cm and 220 cm. A common size for this type of sofa would be around 190 cm or 200 cm. The perspective makes it look quite long. Let's re-examine the floorboards (0:19). Counting more carefully, it seems closer to 11-13 boards. Using 15cm/board: 11*15 = 165, 13*15 = 195. Using 18cm/board: 11*18 = 198, 13*18 = 234. 200 cm seems like a reasonable central estimate.

7.  **Consider Potential Issues:**
    *   **Lens Distortion:** The video uses a wide-angle (fisheye) lens, which distorts perspective, especially near the edges. This makes visual estimation less accurate. Objects closer to the camera appear larger, and straight lines can appear curved.
    *   **Unknown Dimensions:** The exact dimensions of the reference objects (floorboards, furniture) are unknown.
    *   **Viewing Angle:** The camera moves, changing the perspective constantly.

8.  **Formulate the Answer:** Based on the visual estimation using reference objects like the floorboards and the likely size of a typical 3-seater sofa/sofa bed, the largest side (length) appears to be approximately 190-210 cm. I'll state a specific number within this range as the best estimate, acknowledging it's an approximation. 200 cm seems like a plausible and common dimension.

9.  **Final Check:** Does 200 cm (2 meters) look reasonable for the sofa shown in the context of the room? Yes, it fits the visual scale relative to the other furniture and the room size."

---

### Example 5

"Here's my thought process for determining the distance between the table and the shelf:

1.  **Understand the Goal:** The request asks for the distance in meters between the *closest points* of the dining table and the shelf (specifically the bookshelf visible further down the room).

2.  **Analyze the Video:**
    *   **Starting Point (0:00-0:04):** The video starts looking through a doorway with frosted glass panels. We see a long room with a dining table in the foreground and kitchen elements further back. The camera moves forward down a few steps into the room.
    *   **Moving into the Room (0:05-0:14):** The camera pans around the dining area. The dining table is wooden, rectangular, and has several chairs around it. There's a large window on the left wall. On the right wall, there's a fireplace/wood stove. A large black pendant light hangs over the table.
    *   **Focusing on the Kitchen Area (0:15-0:22):** The camera moves further down the room, past the dining table. We see a kitchen island/peninsula with stools. Beyond that is the main kitchen counter area with cabinets, a sink, and windows.
    *   **Panning Back (0:23-0:38):** The camera pans back towards the dining area and the fireplace/wood stove on the right wall. It then continues panning left, showing the stairs leading back up to the initial doorway.
    *   **Further down the room (0:46-0:50):** The camera moves towards the back of the kitchen. We see the main kitchen counter with the stove and range hood. Further down, past the kitchen island, there's a doorway leading into a pantry/room lined with shelves filled with items (jars, cans, etc.). This is the ""shelf"" mentioned in the prompt.
    *   **Moving Around Kitchen (0:51-1:16):** The camera moves around the kitchen island and along the main kitchen counter, eventually looking back towards the dining area and the pantry doorway.

3.  **Identify the Objects:**
    *   **Table:** The prominent wooden dining table in the foreground/middle of the long room.
    *   **Shelf:** The shelving unit visible through the doorway at the far end of the kitchen area (the pantry).

4.  **Determine the ""Closest Points"":**
    *   The closest point of the *table* to the shelf would be the end of the table nearest the kitchen area.
    *   The closest point of the *shelf* to the table would be the front edge of the shelves within the pantry doorway.

5.  **Estimate the Distance:**
    *   **Visualize the Space:** The room is long and relatively narrow. It appears to be an open-plan kitchen/diner.
    *   **Break Down the Distance:**
        *   From the end of the dining table to the *start* of the kitchen island/peninsula. This looks like maybe 2-3 meters of clear floor space.
        *   The length of the kitchen island/peninsula. This looks roughly 1.5-2 meters long.
        *   The space between the kitchen island and the main kitchen counter along the wall. This is the main walkway, perhaps 1-1.5 meters wide. (Though the distance we care about is *along* the length of the room).
        *   From the *end* of the kitchen island (closest to the pantry) to the pantry doorway. This looks like another 2-3 meters.
    *   **Refined Estimation:** Let's trace the path *straight* from the table end to the pantry door.
        *   The table seems to end roughly parallel to the fireplace (0:09).
        *   The kitchen island starts a bit further down (0:15). Let's say 2 meters from table end to island start.
        *   The island itself looks about 1.5 meters deep (front to back, relative to the camera's initial view). Let's estimate its length (along the room) as maybe 2 meters (0:21 shows its length better).
        *   From the end of the island to the pantry doorway (0:48-0:50). This distance looks similar to the gap between the table and the island, maybe another 2-3 meters.
    *   **Total Estimation:** 2m (table to island) + 2m (island length, effectively part of the distance to cover) + 3m (island end to pantry door) = approximately 7 meters.

6.  **Consider Scale Clues:**
    *   **Floorboards:** Standard floorboards are often around 10-15 cm wide. Counting boards is difficult due to perspective and movement, but it suggests a significant length.
    *   **Doors:** Standard door height is ~2m. The pantry doorway looks standard.
    *   **Counter Depth:** Standard kitchen counters are ~60cm deep. The island looks deeper, maybe 90cm-1m.
    *   **Table Length:** A table seating 6-8 people is typically 1.8m to 2.4m long.
    *   **Room Length:** Combining these, the dining area looks maybe 4m long, the kitchen area (island + walkway + counter space before pantry) looks another 4-5m long. Totaling 8-9m from the window end to the pantry door. The table occupies a portion of the dining area.

7.  **Refine based on Scale:** If the whole room (dining + kitchen) is ~8-9m long, and the table takes up maybe 2.5m of that length, starting maybe 1m from the window wall, then the end of the table is roughly 3.5m into the room. The pantry door is at the far end (8-9m). The distance would be 8.5m (total est.) - 3.5m (table end) = 5 meters.

8.  **Reconcile Estimates:** My first estimate was ~7m, my second based on rough room dimensions is ~5m. Let's re-examine the middle ground.
    *   Look at 0:10. The table ends.
    *   Look at 0:15. The camera has moved forward. The kitchen area starts properly.
    *   Look at 0:48. We are near the pantry door.
    *   The distance from the end of the table (around 0:10-0:11) to the pantry doorway (visible clearly around 0:48) seems substantial. The kitchen island occupies a significant chunk of this space. Let's use the island as a reference. It looks around 2m long. The gap *before* the island (from the table) looks similar in length to the gap *after* the island (to the pantry door). Let's say Gap1 = 2m, Island = 2m, Gap2 = 2.5m. Total distance = Gap1 + Island + Gap2 = 2 + 2 + 2.5 = 6.5 meters. This feels more plausible than 5m, given the visual length.

9.  **Final Check:** View the transition from 0:11 (near table end) to 0:48 (near pantry). It's a considerable walk/pan. 6-7 meters seems reasonable for the space shown, accounting for the kitchen island and walkways.

10. **Format the Answer:** State the estimated distance clearly in meters, acknowledging it's an estimate based on visual cues."

---

## Guide for Writing “Thoughts” (Step-by-Step Reasoning for Video QA)

**Purpose:**  
These “thoughts” are written to show a clear, step-by-step reasoning process for answering questions based on video content. They help make the reasoning *transparent*, *logical*, and *explainable*.

---

### Structure of a Good "Thoughts"

1. **Clarify the Question**
   - Start by rephrasing or clearly identifying what the question is asking.
   - Example: *“The goal is to determine which object appears first in the video.”*

2. **Observe the Video and Gather Evidence**
   - Mention specific timestamps (e.g., *“At 00:15, the sink becomes visible”*).
   - Describe what is seen, including spatial positions, sizes, and context.
   - Use clear bullet points or numbered lists if multiple observations are needed.

3. **Use Common Knowledge and Estimation**
   - Use real-world dimensions or typical object sizes for scale comparisons.
   - Example: *“Standard kitchen counters are about 60 cm deep, so the sink appears slightly shorter than that.”*

4. **Compare and Reason**
   - Compare objects, locations, appearances, or distances based on what is seen.
   - Discuss alternatives and edge cases.
   - Example: *“The stool is at the foot of the bed, so it's farther from the toilet than the bed's head, which is near the door.”*

5. **Rule Out Incorrect Options (if applicable)**
   - If a multiple-choice question is given, explain why incorrect choices can be eliminated.
   - Example: *“Option A lists the oven before the table, but the oven appears much later.”*

6. **State a Final Answer**
   - Conclude clearly and confidently.
   - Example: *“Therefore, the correct order is: sofa, table, TV monitor, oven.”*

---

### Style and Tone

- Write in **clear, logical English**.
- Keep the tone **neutral and analytical**.
- Use **present tense** when describing the video content.
- Be detailed but not overly wordy—aim for **clarity over length**.

---

### Bonus Tips

- If uncertain, show multiple interpretations and explain which is more likely.
- Use indentation or line breaks to separate steps and make the reasoning easier to follow.
- Imagine you are teaching someone else *how* to think through the video—be explicit.

---

## Question

**What is the total number of tables shown in this scene?**


Please answer the question above by following the process and format below:

---

## Thoughts

Your reasoning process

---

## Response

Your answer based on the reasoning process

---

## Final Answer

\[
\boxed{Your final answer}
\]

---

\end{Verbatim}
\end{quote}

\vspace{1em}

In the prompt template, there is a placeholder line:

\begin{quote}
  \textit{What is the total number of tables shown in this scene?}
\end{quote}

This placeholder question should be replaced with the actual question corresponding to each video in the dataset.

\twocolumn

\section{Supplementary Material for Section~\ref{sec:additional_probing} (Additional Probing Experiments)}

The figures and tables supplementing Section~\ref{sec:additional_probing} (Additional Probing Experiments) are provided in this chapter/section.

\begin{table}[H]
  \small
  \centering
  \setlength{\tabcolsep}{4pt}
  \begin{tabular}{lcc|c}
    \toprule
    & \textbf{ViCA-Base-7B} & \textbf{ViCA-7B} & \textbf{ViCA-7B} \\
    & w/o time\_inst. & w/o time\_inst. & w/ time\_inst. \\
    \midrule
    \rowcolor{gray!20} \multicolumn{4}{l}{\textit{Numerical Answer}} \\
    Obj. Count         & 65.6 & \textbf{68.8} & 68.4 (-0.4) \\
    Abs. Dist.         & 51.3 & \textbf{57.0} & 56.9 (-0.1) \\
    Obj. Size          & 74.9 & 79.2 & \textbf{79.3} \\
    Room Size          & 67.1 & 75.1 & \textbf{77.6} \\
    \midrule
    \rowcolor{gray!20} \multicolumn{4}{l}{\textit{Multiple-Choice Answer}} \\
    Rel. Dist.         & 52.0 & \textbf{58.5} & 58.0 (-0.5) \\
    Rel. Dir.          & 32.6 & \textbf{42.6} & 40.8 (-1.8) \\
    Route Plan         & 28.4 & \textbf{34.5} & 28.9 (-5.6) \\
    Appr. Order        & \textbf{70.9} & 68.8 & 69.1 (-1.8) \\
    \midrule
    \textbf{Average}   & 55.4 & \textbf{60.6} & 59.9 (-0.7) \\
    \bottomrule
  \end{tabular}
  \caption{
  Comparison of ViCA variants with and without time instruction.
  The third column (ViCA-7B w/ time\_instruction) is compared against the better of the two baselines (ViCA-Base-7B and ViCA-7B w/o time\_instruction).
  The values in parentheses indicate the performance difference from the best baseline for each metric.
  Bold highlights denote the best overall score among the three models.
  }
  \label{tab:vica-time-instruction}
  \end{table}
  
  \begin{table}[H]
  \small
  \centering
  \setlength{\tabcolsep}{4pt}
  \begin{tabular}{lcc|c}
    \toprule
    & \textbf{ViCA-Base-7B} & \textbf{ViCA-7B} & \textbf{ViCA-7B} \\
    & 64 frames & 64 frames & 128 frames \\
    \midrule
    \rowcolor{gray!20} \multicolumn{4}{l}{\textit{Numerical Answer}} \\
    Obj. Count         & 65.6 & \textbf{68.8} & 68.1 (-0.7) \\
    Abs. Dist.         & 51.3 & \textbf{57.0} & 54.2 (-2.8) \\
    Obj. Size          & 74.9 & 79.2 & \textbf{79.3} \\
    Room Size          & 67.1 & \textbf{75.1} & 74.5 (-0.7) \\
    \midrule
    \rowcolor{gray!20} \multicolumn{4}{l}{\textit{Multiple-Choice Answer}} \\
    Rel. Dist.         & 52.0 & \textbf{58.5} & 58.3 (-0.1) \\
    Rel. Dir.          & 32.6 & \textbf{42.6} & 40.4 (-2.2) \\
    Route Plan         & 28.4 & \textbf{34.5} & 30.4 (-4.1) \\
    Appr. Order        & \textbf{70.9} & 68.8 & 63.1 (-7.8) \\
    \midrule
    \textbf{Average}   & 55.4 & \textbf{60.6} & 58.5 (-2.0) \\
    \bottomrule
  \end{tabular}
  \caption{
  Comparison of ViCA variants across different frame settings.
  The third column (ViCA-7B 128 frames) is compared against the better of the two baselines (ViCA-Base-7B and ViCA-7B with 64 frames).
  The values in parentheses indicate the performance difference from the best baseline for each metric.
  Bold highlights denote the best overall score among the three models.
  }
  \label{tab:vica-frame-comparison}
  \end{table}

\section{\textit{Thought-Response-Final Answer} Examples generated by ViCA-7B-Thinking}

In this section, we present several examples illustrating the \textit{Thought-Response-Final Answer} structure produced by our ViCA-7B-Thinking model.

\label{app:thinking_example}

\end{document}